\documentclass[10pt,journal,compsoc]{IEEEtran}

\usepackage{filecontents,lipsum}
\usepackage[noadjust]{cite}
\usepackage{xcolor}
\usepackage{dsfont}
\newcommand{\etal}{\textit{et al}.}

\ifCLASSINFOpdf
  \usepackage[pdftex]{graphicx}
\else
\fi
\usepackage{amsmath}
\usepackage{acronym}
\usepackage{url}

\ifCLASSINFOpdf
 \usepackage[pdftex]{thumbpdf}
\else
 \usepackage[dvips]{thumbpdf}
\fi

\usepackage{hyperref}

\usepackage{textcomp}

\begin{document}
%
\title{Grad-CAM++: Improved Visual Explanations for Deep Convolutional Networks}
%
%
%
%

\author{Aditya~Chattopadhyay\IEEEauthorrefmark{1},
        Anirban Sarkar\IEEEauthorrefmark{1},~\IEEEmembership{Member,~IEEE,}
        Prantik Howlader,
   and~Vineeth~N~Balasubramanian,~\IEEEmembership{Member,~IEEE}

\IEEEcompsocitemizethanks{\IEEEcompsocthanksitem\IEEEauthorrefmark{1}These two authors contributed equally to the work.\protect
\IEEEcompsocthanksitem A. Chattopadhyay, A. Sarkar and V. Balasubramanian are with the Department of Computer Science and Engineering, Indian Institute of Technology Hyderabad,
Telangana, India.\protect\\
E-mail IDs: \{adityac, cs16resch11006, vineethnb\}@iith.ac.in
\IEEEcompsocthanksitem P. Howlader is with Cisco Systems, Bangalore, India.\protect

}
}

%
%

\markboth{Transactions on Pattern Analysis and Machine Intelligence}%
{Shell \MakeLowercase{\textit{et al.}}: Bare Advanced Demo of IEEEtran.cls for IEEE Computer Society Journals}
%



\IEEEtitleabstractindextext{%
\begin{abstract}
Over the last decade, Convolutional Neural Network (CNN) models have been highly successful in solving complex vision problems. However, these deep models are perceived as "black box" methods considering the lack of understanding of their internal functioning. There has been a significant recent interest in developing explainable deep learning models, and this paper is an effort in this direction. Building on a recently proposed method called Grad-CAM, we propose a generalized method called Grad-CAM++ that can provide better visual explanations of CNN model predictions, in terms of better object localization as well as explaining occurrences of multiple object instances in a single image, when compared to state-of-the-art. We provide a mathematical derivation for the proposed method, which uses a weighted combination of the positive partial derivatives of the last convolutional layer feature maps with respect to a specific class score as weights to generate a visual explanation for the corresponding class label. Our extensive experiments and evaluations, both subjective and objective, on standard datasets showed that Grad-CAM++ provides promising human-interpretable visual explanations for a given CNN architecture across multiple tasks including classification, image caption generation and 3D action recognition; as well as in new settings such as knowledge distillation.  
\end{abstract}

}

\maketitle

\pagestyle{empty}
\thispagestyle{empty}

\IEEEdisplaynontitleabstractindextext

%
\IEEEpeerreviewmaketitle

\ifCLASSOPTIONcompsoc
\IEEEraisesectionheading{\section{Introduction}\label{sec:introduction}}
\else
\section{Introduction}
\label{sec:introduction}
\fi

%
%
%
%
The dramatic progress of machine learning in the form of deep neural networks
has opened up new Artificial Intelligence (AI) capabilities in real-world applications. It is no new
fact that deep learning models offer tremendous benefits with impressive results in tasks like
object detection, speech recognition, machine translation to name a few. However, the connectionist approach of deep learning is fundamentally different from earlier AI systems where the predominant reasoning methods were logical and symbolic. These early systems could generate a trace of their inference steps, which then became the basis for explanation. On the other hand, the effectiveness of today's intelligent systems is limited by the inability to explain their decisions to human users. This issue is especially important for risk-sensitive applications such as security, clinical decision support or autonomous navigation.



To this end, various methods have been proposed by researchers over the last few years to know what is beneath the hood when using deep learning models. For instance, one category of methods rationalize/justify the decision of a model by training another deep model which comes up with explanations as to why the model behaved the way it did. Another approach has been to probe the black-box neural network models by trying to change the input intelligently and analyzing the model's response to it. While there has been promising progress in this area (a detailed survey is presented in Section \ref{Section:Related work}), existing efforts are limited and the objective to achieve explainable deep learning has a long way to go, considering the difficulty and variations in problem scope. 

In the context of understanding Convolutional Neural Networks (CNNs), Zeiler \& Fergus \cite{Zeiler} made one of the first efforts in understanding what a CNN learns. Their method, however, involves significant computations to generate this understanding. Zhou \etal followed up on the same objective in \cite{Zhou} and showed that various layers of the CNN behave as unsupervised object detectors using a new technique called CAM (Class Activation Mapping). By using a global average pooling \cite{Lin} layer, and visualizing the
weighted combination of the resulting feature maps at the penultimate (pre-softmax) layer, they were able to obtain heat maps that explain which parts of an input image were looked at by the CNN for assigning a label. However, this technique involved retraining a linear classifier for each class. Similar methods were examined with different pooling layers such as global max pooling in \cite{Oquab} and log-sum-exp pooling in \cite{Pinheiro}. Selvaraju \etal subsequently \cite{Selvaraju} came up with an efficient generalization of CAM, known as Grad-CAM, which fuses the class-conditional property of CAM with existing pixel-space gradient visualization techniques such as Guided Back-propagation \cite{Springenberg} and Deconvolution \cite{Zeiler} to highlight fine-grained details on the image. Grad-CAM made CNN-based models more transparent by visualizing input regions with high resolution details that are important for predictions. 

While the visualizations generated by gradient-based methods such as Grad-CAM provide explanations for the prediction made by the CNN model with fine-grained details of the predicted class, these methods have limitations - for example, their performance drops when localizing multiple occurrences of the same class (Figure \ref{fig:Grad-CAM++ improvements}). In addition, for single object images, Grad-CAM heatmaps often do not capture the entire object in completeness, which we show is required for better performance on the associated recognition task. To address these limitations and to extend the visualizations to spatiotemporal data such as videos, in this work, we propose Grad-CAM++, a generalized visualization technique for explaining CNN decisions, which ameliorates the aforementioned flaws and provides a more general approach. Our key contributions in this work are summarized as follows:

\begin{itemize}
\item We introduce pixel-wise weighting of the gradients of the output w.r.t. a particular spatial position in the final convolutional feature map of the CNN. This approach provides a measure of importance of each pixel in a feature map towards the overall decision of the CNN. Importantly, we derive closed-form solutions for the pixel-wise weights, as well as obtain exact expressions for higher order derivatives for both softmax and exponential output activation functions. Our approach requires a single backward pass on the computational graph, thus making it computationally equivalent to prior gradient-based methods while giving better visualizations.
\item While several methods exist to visualize CNN decisions, namely, Deconvolution, Guided Back-propagation, CAM, and Grad-CAM, the assessment of the quality of visualizations is done mainly through human evaluations or some auxiliary metric like localization error with respect to bounding boxes (ground truth). 
This need not correlate with the actual factors responsible for the network's decision. We propose new metrics in this work to evaluate (objectively) the faithfulness of the proposed explanations to the underlying model, i.e., whether the visualization directly correlates with the decision. Our results with these metrics show  superior performance of Grad-CAM++ over state-of-the-art.
\item In accordance with previous efforts in visualizing CNNs, we also conduct human studies to test the quality of our explanations. These studies show that the visualizations produced by Grad-CAM++ instill greater trust in the underlying model (for the human user) than the corresponding visualizations produced by Grad-CAM. 
\item Through both visual examples and objective evaluations, we also show that Grad-CAM++ improves upon Grad-CAM in weakly supervised localization of object classes in a given image.
\item A good explanation should be able to effectively distill its knowledge. This aspect of explainable-AI is largely ignored in recent works. We show that in a constrained teacher-student setting, it is possible to achieve an improvement in the performance of the student by using a specific loss function inspired from the explanation maps generated by Grad-CAM++. We introduce a training methodology towards this objective, and show promising results of students trained using our methodology.
\item Lastly, we show the effectiveness of Grad-CAM++ in other tasks (beyond recognition) - in particular, image captioning and 3D action recognition. Visualization of CNN decisions so far have largely been limited to 2D image data, and this is one of very few efforts (one similar recent effort is \cite{bargal2018excitation}) on \textit{visual explanations} of 3D-CNNs in video understanding .
\end{itemize}



\section{Related Work}
\label{Section:Related work}

In this section, we present a survey of related efforts in understanding the predictions of CNNs in recent years.
\begin{figure*}[t]
\begin{center}
   \includegraphics[scale=1.3]{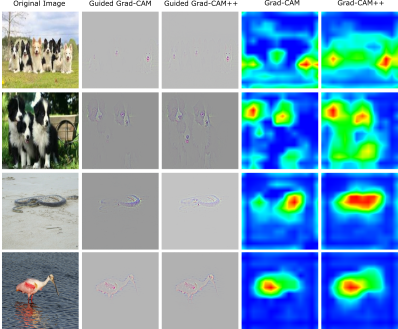}
\end{center}
   \caption{Success of Grad-CAM++ for: (a) multiple occurrences of the same class (Rows 1-2), and (b) localization capability of an object in an image (Rows 3-4). Note: All dogs are better visualized in the Grad-CAM++ and Guided Grad-CAM++ saliency maps for input images of rows 1 and 2 as compared to Grad-CAM. Similarly, the entire region of the class is localized for input images of rows 3 and 4 (full body of the snake and the head/legs of the bird). Grad-CAM heatmaps only exhibit partial coverage.}
\label{fig:Grad-CAM++ improvements}
\vspace{-5mm}\end{figure*}
As mentioned earlier, Zeiler \& Fergus \cite{Zeiler} proposed one of the first efforts in this area of understanding deep CNNs, and developed a deconvolution approach to better understand what the higher layers in a given network have learned. ``Deconvnet" makes data flow from a neuron activation in the higher layers, down to the image. In this process, parts of the image that strongly activate that neuron get highlighted. Springenberg \etal \cite{Springenberg} extended this work to \textit{guided backpropagation} which helped understand the impact of each neuron in a deep network w.r.t. the input image. These visualization techniques were compared in \cite{Mahendran}. Yosinski \etal \cite{jason} proposed a method to synthesize the input image that causes a specific unit in a neural network to have a high activation, for visualizing the functionality of the unit. A more guided approach to synthesizing input images that maximally activate a neuron was proposed by Simonyan \etal in \cite{Simonyan}. In this work, they generated class-specific saliency maps by performing a gradient ascent in pixel space to reach a maxima. This synthesized image serves as a class-specific visualization and helps understand how a given CNN modeled a class.

From a different perspective, Ribeiro \etal \cite{Ribeiro} introduced LIME (Local Interpretable Model-Agnostic Explanations), a method which makes a local
approximation to the complex decision surface of any deep model with simpler interpretable classifiers like sparse linear models or shallow decision trees. For every test point, analyzing the weights of the sparse linear model gives an intuition to a non-expert as to the relevance of that feature in that particular prediction. Shrikumar \etal\cite{shrikumar2017learning} more recently proposed DeepLift  which evaluates the importance of each input neuron for a particular decision by approximating the instantaneous gradients (of the output with respect to the inputs) with discrete gradients. This obviates the need to train interpretable classifiers for explaining each input-output relations (as in LIME) for every test point. In another approach, Al-Shedivat \etal \cite{Al-Shedivat} proposed Contextual Explanation Networks (CENs), a class of models that jointly learns to predict and explain its decision. Unlike existing posthoc model-explanation tools, CENs combine deep networks with context-specific probabilistic models and construct explanations in the form of locally-correct hypotheses. Konam \cite{Konam} developed an algorithm to detect specific neurons which are responsible for decisions taken by a network and additionally locate patches of an input image which maximally activate those neurons. Lengerich \etal \cite{Lengerich} proposed a different method altogether, where instead of explaining the decision in terms of the input, they developed statistical metrics to evaluate the relation between the hidden representations in a network and its prediction. Another recent work \cite{Kim}, focusing on interpretability for self-driving cars, trained a visual attention model followed by a CNN model to obtain potentially salient image regions and applied causal filtering to find true input regions that actually influence the output.
	
In spite of these recent developments, we are still far from the desired goal of interpretable deep learning models, and there is a continued need to develop algorithms that can generate interpretable explanations of the results of deep models used across domains. A key objective of these efforts is to build trust in these systems when integrating them into our daily lives. Our work in this paper is mainly inspired by two algorithms, namely CAM \cite{Zhou} and Grad-CAM \cite{Selvaraju}, which are widely used today \cite{Li2018}. In CAM, the authors demonstrate that a CNN with a Global Average Pooling (GAP) layer shows localization capabilities despite not being explicitly trained to do so. In a CNN with GAP, the final classification score  $Y^c$ for a particular class $c$ can be written as a linear combination of its global average pooled last convolutional layer feature maps $A^k$. 
\begin{equation}
 Y^c = \sum_{k}w^c_k.\sum_{i}\sum_{j}A_{ij}^k\label{fundamental_eq}
\end{equation}
\noindent Each spatial location $(i,j)$ in the class-specific saliency map $L^c$ is then calculated as: 
\begin{equation}
 L_{ij}^c = \sum_{k}w^c_k.A_{ij}^k
 \label{saliency_map_generic} 
\end{equation}
$L^c_{ij}$ directly correlates with the importance of a particular spatial location $(i,j)$ for a particular class $c$ and thus functions as a visual explanation of the class predicted by the network. CAM estimates these weights $w_k^c$ by training a linear classifier for each class $c$ using the activation maps of the last convolutional layer generated for a given image. This however limits its explainability prowess to CNNs with a GAP penultimate layer, and requires retraining of multiple linear classifiers (one for each class), after training of the initial model.

Grad-CAM was built to address these issues. This approach \cite{Selvaraju} defines the weights $w_k^c$ for a particular feature map $A^k$ and class $c$ as:
\begin{equation}
 w_k^c = \frac{1}{Z}\sum_{i}\sum_{j}\frac{\partial Y^c}{\partial A_{ij}^k}\label{global_average_wt.}
\end{equation}
where $Z$ is a constant (number of pixels in the activation map). Grad-CAM can thus work with any deep CNN where the final $Y^c$ is a differentiable function of the activation maps $A^k$, without any retraining or architectural modification. To obtain fine-grained pixel-scale representations, the Grad-CAM saliency maps are upsampled and fused via point-wise multiplication with the visualizations generated by Guided Backpropagation \cite{Springenberg}. This visualization is referred to as Guided Grad-CAM. 

This approach however has some shortcomings as shown in Fig \ref{fig:Grad-CAM++ improvements}. Grad-CAM fails to properly localize objects in an image if the image contains multiple occurrences of the same class. This is a serious issue as multiple occurrences of the same object in an image is a very common occurrence in the real world. Another consequence of an unweighted average of partial derivatives is that often, the localization doesn't correspond to the entire object, but bits and parts of it. This can hamper the user's trust in the model, and impede Grad-CAM's premise of making a deep CNN more transparent. 

In this work, we propose a generalization to Grad-CAM which addresses the abovementioned issues and consequently serves as a better explanation algorithm for a given CNN architecture, and hence the name for the proposed method, Grad-CAM++. We derive closed-form solutions for the proposed method and carefully design experiments to evaluate the competence of Grad-CAM++ both objectively and subjectively. In all our experiments, we compare the performance of our method with Grad-CAM as it is considered the current state-of-the-art CNN discriminative (class specific saliency maps) visualization technique \cite{Li2018}. We now present the proposed methodology, beginning with its intuition.


\section{Grad-CAM++: Proposed Methodology} 
\subsection{Intuition}
\label{Section:Intuition}

\begin{figure*}[t]
\begin{center}
   \includegraphics[scale=0.50
   ]{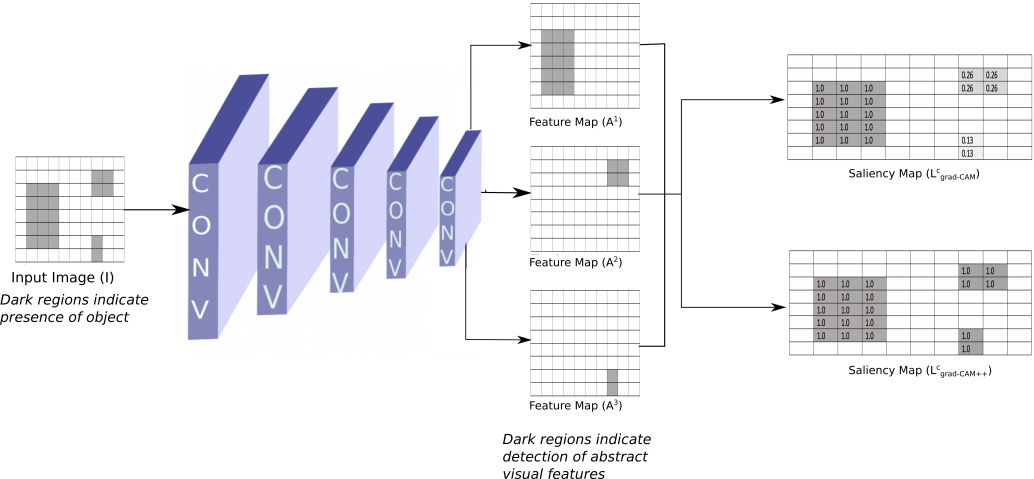}
\end{center}
   \caption{A hypothetical example elucidating the intuition behind grad-CAM++. The CNN task here is binary object classification. Clearly taking a weighted combination of gradients $L^c_{grad-CAM++}$ provides better salient features (all the spatially relevant regions of the input image are equally highlighted) than its unweighted counterpart $L^c_{grad-CAM}$ (some parts of the object are paled out in the saliency map). The values in the pixels of each saliency map indicates the intensity at that point.}
\label{fig:intuition}
\vspace{-5mm}\end{figure*}

Consider a saliency map $L^c$ (as defined in Eqn \ref{saliency_map_generic}  where $i$ \& $j$ are the iterators over the pixels in the map), and a binary object classification task, with output $0$ if object is absent or $1$ if present. (For image $I$ in Fig. \ref{fig:intuition}, the network outputs $1$.) $A^k$ represents the visualization of the $k^{th}$ feature map. According to previous work \cite{zhou2014object, Zeiler}, each $A^k$ is triggered by an abstract visual pattern. In this example, $A^k_{ij}= 1$ if a visual pattern is detected else $0$. (In Fig. \ref{fig:intuition}, the dark regions correspond to $A^k_{ij}= 1$.) The derivative $\frac{\partial y^c}{\partial A_{ij}^k}$ is expected to be high for feature map pixels that contribute to the presence of the object. Without loss of generality, let us assume the derivative map to be: 
\begin{align}
  \frac{\partial y^c}{\partial A_{ij}^k} = 1 \qquad if \quad A^k_{ij} = 1 \nonumber \\
                                        = 0 \qquad if \quad A^k_{ij} = 0 \label{eq:toy derivative map} 
\end{align}
Plugging in values from Eqn \ref{eq:toy derivative map} into Eqn \ref{global_average_wt.}, we obtain the following feature map weights in the case of Grad-CAM for the given input image $I$, $w_1^c = \frac{15}{80}$, $w_2^c = \frac{4}{80}$ and $w_3^c = \frac{2}{80}$ for the three feature maps. Here $Z = 80$, the number of pixels in the feature map. The Grad-CAM saliency map $L^c_{grad-CAM}$ is thus obtained using Eqn \ref{saliency_map_generic} (refer Fig. \ref{fig:intuition}). 
Comparing with the input image $I$, it is evident that the spatial footprint of an object in an image is important for Grad-CAM's visualizations to be strong. Hence, if there were multiple occurrences of an object with slightly different orientations or views (or parts of an object that excite different feature maps), different feature maps may be activated with differing spatial footprints, and the feature maps with lesser footprints fade away in the final saliency map.

This problem can be fixed by taking a weighted average of the pixel-wise gradients. In particular, we reformulate Eqn \ref{global_average_wt.} by explicitly coding the structure of the weights $w_k^c$ as: 
\begin{equation}
 w_k^c = \sum_{i}\sum_{j}\alpha_{ij}^{kc}.relu(\frac{\partial Y^c}{\partial A_{ij}^k}) \label{weighted version}
\end{equation}

\noindent where \textit{relu} is the Rectified Linear Unit activation function. Here the $\alpha_{ij}^{kc}$'s are weighting co-efficients for the pixel-wise gradients for class $c$ and convolutional feature map $A^k$. In the above example, by taking 
\begin{align}
   \alpha_{ij}^{kc} = \frac{1}{\sum_{l,m}\frac{\partial y^c}{\partial A_{lm}^k}} \qquad if \quad \frac{\partial y^c}{\partial A_{ij}^k} = 1 \nonumber \\
    = 0                              \qquad \text{otherwise} \label{eq:alphas in toy example} 
\end{align}
\noindent presence of objects in all feature maps are highlighted with equal importance.

The idea behind considering only the positive gradients in Eqn \ref{weighted version} is similar to works such as Deconvolution \cite{Zeiler} and Guided Backpropogation \cite{Springenberg}. $w_k^c$ captures the importance of a particular activation map $A^k$, and we prefer positive gradients to indicate visual features that increase the output neuron's activation, rather than suppress the output neuron's activation. An empirical verification of this ``positive gradients'' hypothesis is presented later in Section \ref{subsec_why_positive_grads}. 

We now present the proposed methodology.

\subsection{Methodology} 
\label{Section:Approach}

\begin{figure}[t]
\begin{center}
   \includegraphics[scale=0.13
   ]{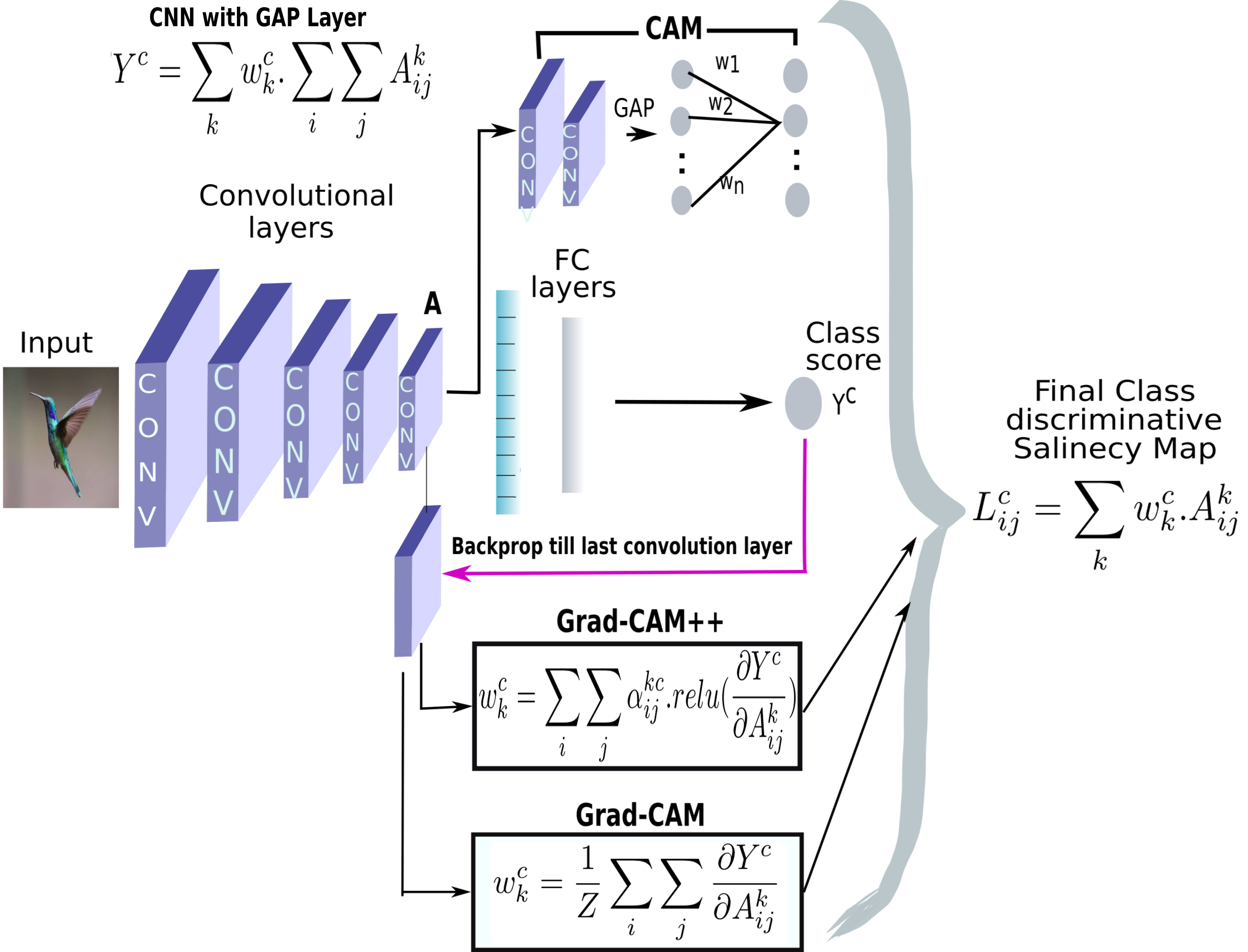}
\end{center}
   \caption{An overview of all the three methods -- CAM, Grad-CAM, Grad-CAM++ -- with their respective computation expressions.}
\label{fig:eagle's eye view}
\vspace{-5mm}\end{figure}
We derive a method for obtaining the gradient weights $\alpha_{ij}^{kc}$ for a particular class $c$ and activation map $k$. Let $Y^c$ be the score of a particular class $c$. Combining Eqn \ref{fundamental_eq} and Eqn \ref{weighted version}, we get:
\begin{equation}
 Y^c = \sum_{k}\{\sum_{a}\sum_{b}\alpha_{ab}^{kc}.relu(\frac{\partial Y^c}{\partial A_{ab}^k})\}[\sum_{i}\sum_{j} A_{ij}^k] \\
\end{equation}
Here, $(i,j)$ and $(a,b)$ are iterators over the same activation map $A^k$ and are used to avoid confusion. Without loss of generality, we drop the $relu$ in our derivation as it only functions as a threshold for allowing the gradients to flow back.
Taking partial derivative w.r.t. $A_{ij}^k$ on both sides: 
\begin{equation}
 \frac{\partial Y^c}{\partial A_{ij}^k} = \sum_{a}\sum_{b}\alpha_{ab}^{kc}.\frac{\partial Y^c}{\partial A_{ab}^k}\! + \! \sum_{a}\sum_{b} A_{ab}^k\{\alpha_{ij}^{kc}.\frac{\partial ^2Y^c}{(\partial A_{ij}^k)^2}\}
\end{equation}
\noindent Taking a further partial derivative w.r.t. $A_{ij}^k$:
\begin{equation}
 \frac{\partial ^2Y^c}{(\partial A_{ij}^k)^2} = 2.\alpha_{ij}^{kc}.\frac{\partial ^2Y^c}{(\partial A_{ij}^k)^2}\! + \! \sum_{a}\sum_{b} A_{ab}^k\{\alpha_{ij}^{kc}.\frac{\partial ^3Y^c}{(\partial A_{ij}^k)^3}\}
\end{equation}
\noindent Rearranging terms, we get:
\begin{equation}
\label{eq_final_alpha}
 \alpha_{ij}^{kc} = \frac{\frac{\partial ^2Y^c}{(\partial A_{ij}^k)^2}}{2 \frac{\partial ^2Y^c}{(\partial A_{ij}^k)^2}\! + \! \sum_{a}\sum_{b} A_{ab}^k\{\frac{\partial ^3Y^c}{(\partial A_{ij}^k)^3}\}}
\end{equation}
\noindent Substituting Eqn \ref{eq_final_alpha} in Eqn \ref{weighted version}, we get the following Grad-CAM++ weights:
\begin{equation}
 w^c_k = \sum_{i}\sum_{j}[\frac{\frac{\partial ^2Y^c}{(\partial A_{ij}^k)^2}}{2 \frac{\partial ^2Y^c}{(\partial A_{ij}^k)^2}\! + \! \sum_{a}\sum_{b} A_{ab}^k\{\frac{\partial ^3Y^c}{(\partial A_{ij}^k)^3}\}}].relu(\frac{\partial Y^c}{\partial A_{ij}^k})
\end{equation}
Evidently, comparing with Eq \ref{global_average_wt.}, if $\forall i,j, \quad \alpha_{ij}^{kc} = \frac{1}{Z}$, Grad-CAM++ reduces to the formulation for Grad-CAM. Thus, Grad-CAM++, as its name suggests, can be (loosely) considered a generalized formulation of Grad-CAM.

In principle, the class score $Y^c$ can be any prediction; the only constraint being that $Y^c$ must be a smooth function. For this reason, unlike Grad-CAM (which takes the penultimate layer representation as their class score $Y^c$), we pass the penultimate layer scores through an exponential function, as the exponential function is infinitely differentiable.

In Fig \ref{fig:Grad-CAM++ improvements}, we illustrate visually the advantage of Grad-CAM++ over Grad-CAM. A bird's eye view of all the three methods -- CAM, Grad-CAM, and Grad-CAM++ -- is presented in Fig. \ref{fig:eagle's eye view}.

\subsection{Computation Analysis}
The time overhead for calculating higher-order derivatives remains of the same order as Grad-CAM, as only the diagonal terms are used (no cross higher-order derivatives). If we pass the penultimate layer scores through an exponential function and the last layer has only linear or ReLU activation functions, the calculation of higher-order derivatives becomes trivial. Let $S^c$ be the penultimate layer scores for class $c$. 
\begin{equation}
	Y^c = \exp(S^c)
\end{equation}
\begin{equation}
	\frac{\partial Y^c}{\partial A_{ij}^k} = \exp(S^c)\frac{\partial S^c}{\partial A_{ij}^k}
\end{equation}
The quantity $\frac{\partial S^c}{\partial A_{ij}^k}$ can be easily calculated using machine learning libraries such as PyTorch or TensorFlow, which implement automatic differentiation.
\begin{equation}
	\frac{\partial ^2Y^c}{(\partial A_{ij}^k)^2} = \exp(S^c)\bigg[\bigg(\frac{\partial S^c}{(\partial A_{ij}^k)}\bigg)^2 + \frac{\partial ^2S^c}{(\partial A_{ij}^k)^2}\bigg]
    \label{double-derivative relu}
\end{equation}
\noindent Now, assuming a ReLU activation function, $f(x) = \max(x,0)$, its derivatives are given by:


\begin{equation}
\begin{split}
 \frac{\partial f(x)}{\partial x} & = 1 \quad x > 0 \\
 	& = 0 \quad x \leq 0
\end{split}
\end{equation}

\begin{equation}
	\frac{\partial^2 f(x)}{\partial x^2} = 0
    \label{for relu double derivative 0 eq}
\end{equation}

\noindent Eq \ref{for relu double derivative 0 eq} holds even if the activation function is linear. Inserting Eq \ref{for relu double derivative 0 eq} into Eqn \ref{double-derivative relu}, we have:
\begin{equation}
	\frac{\partial ^2Y^c}{(\partial A_{ij}^k)^2} = \exp(S^c)\bigg(\frac{\partial S^c}{\partial A_{ij}^k)}\bigg)^2
    \label{double-derivative answer}
\end{equation}
\noindent Similarly,
\begin{equation}
	\frac{\partial ^3Y^c}{(\partial A_{ij}^k)^3} = \exp(S^c)\bigg(\frac{\partial S^c}{\partial A_{ij}^k}\bigg)^3
    \label{triple-derivative answer}
\end{equation}
\noindent Inserting Eqn \ref{double-derivative answer} and Eqn \ref{triple-derivative answer} into Eqn \ref{eq_final_alpha}, we get:
\begin{equation}
 \alpha_{ij}^{kc} = \frac{(\frac{\partial S^c}{\partial A_{ij}^k})^2}{2 (\frac{\partial S^c}{\partial A_{ij}^k})^2\! + \! \sum_{a}\sum_{b} A_{ab}^k (\frac{\partial S^c}{\partial A_{ij}^k })^3}
\end{equation}
\noindent With a single backward pass on the computational graph, all the gradient weights $\alpha_{ij}^{kc}$ (as defined in Eqn \ref{weighted version}) can be computed. We used the exponential function due to its simplicity. Other smooth functions such as the softmax activation function can also be used with corresponding closed-form expressions to compute the weights. The derivation of the gradient weights for softmax is given below in Section \ref{subsec_softmax_gradient_wts}.

The saliency maps for a given image, $L^c$ is then calculated as a linear combination of the forward activation maps, followed by a $relu$ layer: 
\begin{equation}
	L_{ij}^c = relu(\sum_{k}w^c_k.A_{ij}^k)
    \label{saliency_maps}
\end{equation}
Similar to Grad-CAM, to generate the final saliency maps, we carry out pointwise multiplication of the upsampled (to image resolution) saliency map $L^c$ with the pixel-space visualization generated by Guided Backpropagation. The representations thus generated are hence called \textit{Guided Grad-CAM++}. 

\subsection{Gradient Weights for Softmax Function}
\label{subsec_softmax_gradient_wts}
Like the exponential function, the softmax function is also smooth and commonly used to obtain final class probabilities in classification scenarios. In this case, the final class score $Y^c$ is:
\begin{equation}
Y^c = \frac{exp(S^c)}{\Sigma_k exp(S^k)}
\end{equation}
where the index $k$ runs over all output classes and $S^k$ is the score pertaining to output class $k$ in the penultimate layer. 
\begin{equation}
\frac{\partial Y^c}{\partial A_{ij}^k} = Y^c\bigg[\frac{\partial S^c}{\partial A_{ij}^k} - \Sigma_k Y^k \frac{\partial S^k}{\partial A_{ij}^k}\bigg]
\end{equation}
If the neural network has just linear or ReLU activation functions then $\frac{\partial^2 S^c}{(\partial A_{ij}^k)^2}$ would be 0 (Eqn \ref{for relu double derivative 0 eq}).
\begin{multline}
\frac{\partial^2 Y^c}{(\partial A_{ij}^k)^2} = \frac{\partial Y^c}{\partial A_{ij}^k}\bigg[\frac{\partial S^c}{\partial A_{ij}^k} - \Sigma_k Y^k \frac{\partial S^k}{\partial A_{ij}^k}\bigg]\\ - Y^c\bigg(\Sigma_k \frac{\partial Y^k}{\partial A_{ij}^k} \frac{\partial S^k}{\partial A_{ij}^k}\bigg)
\label{double derivative softmax}
\end{multline}

\begin{multline}
\frac{\partial^3 Y^c}{(\partial A_{ij}^k)^3} = \frac{\partial^2 Y^c}{(\partial A_{ij}^k)^2}\bigg[\frac{\partial S^c}{\partial A_{ij}^k} - \Sigma_k Y^k \frac{\partial S^k}{\partial A_{ij}^k}\bigg]\\ - 2\frac{\partial Y^c}{\partial A_{ij}^k}\bigg(\Sigma_k \frac{\partial Y^k}{\partial A_{ij}^k} \frac{\partial S^k}{\partial A_{ij}^k}\bigg) - Y^c\bigg(\Sigma_k \frac{\partial^2 Y^k}{(\partial A_{ij}^k)^2} \frac{\partial S^k}{\partial A_{ij}^k}\bigg)
\label{triple derivative softmax}
\end{multline}

Plugging Eqn \ref{double derivative softmax} and Eqn \ref{triple derivative softmax} in Eqn \ref{eq_final_alpha}, we get the gradient weights. Note that although evaluating the gradient weights in the case of the softmax function is more involved than the case of the exponential function, it can still be computed via a single backward pass on the computation graph for computing the $\frac{\partial S^k}{\partial A_{ij}^k}$ terms.


\section{Experiments and Results}
\label{Section:Empirical evaluation of generated explanations}
We conducted a comprehensive set of experiments to study the correlation of the visual explanation  with the model prediction (faithfulness) as well as human interpretability (trust). Our experiments involved both objective and subjective assessment, as presented in this section. For all experiments, we used an off-the-shelf VGG-16 \cite{Simonyan1} model from the Caffe Model Zoo \cite{Jia}, to be consistent with earlier work that used the same model \cite{Selvaraju}. We also show results with AlexNet \cite{krizhevsky2012imagenet} and ResNet-50 \cite{he2016deep} architectures in the Appendix. The implementation of our method is publicly available at \href{https://github.com/adityac94/Grad_CAM_plus_plus}{https://github.com/adityac94/Grad\_CAM\_plus\_plus}.

\begin{figure}[t]
\begin{center}
   \includegraphics[scale=1.75]{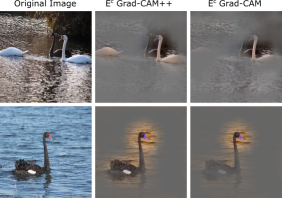}
\end{center}
   \caption{Example explanation maps for 2 images generated by Grad-CAM++ and Grad-CAM.}
\label{fig:explanation map}
\vspace{-5mm}\end{figure}
\subsection{Objective Evaluation for Object Recognition}
\label{Section:Objective evaluation of performance based on object recognition}
We first evaluate the faithfulness of the explanations generated by Grad-CAM++ for the object recognition task. For every image, a corresponding explanation map $E^c$ is generated by point-wise multiplication of the class-conditional saliency maps (upsampled to image resolution) with the original image: 
\begin{equation}
	E^c = L^c \circ I
    \label{explaination_maps}
\end{equation}
where $\circ$ refers to the Hadamard product, $I$ is the input image, $c$ is the class label predicted by the model, and $L^c$ is the class-conditional saliency maps as in Eqn \ref{saliency_maps}. The same procedure was followed for both Grad-CAM++ and Grad-CAM. Sample qualitative results are shown in Fig. \ref{fig:explanation map}. 
We also studied the performance with three different metrics: (i) Average drop \%; (ii) \% increase in confidence; and (iii) Win \% - each of which is described below.\\

\noindent \textbf{(i) Average Drop \%:} 
A good explanation map for a class should highlight the regions that are most relevant for decision-making. It is expected that removing parts of an image will reduce the confidence of the model in its decision, as compared to its confidence when the full image is provided as input. We leverage this to study the performance of the explanation maps generated by Grad-CAM++ and Grad-CAM. We note the change in confidence between the setting when the full image is provided as input, and the setting when only the explanation maps are provided as input. While a reduction in confidence is expected for both Grad-CAM++ and Grad-CAM (possibly due to removal of context), we hypothesize that Grad-CAM++ maintains a higher confidence in the predicted (correct) label than Grad-CAM. This suggests that the visual explanation of Grad-CAM++ includes more of what is relevant (be it the object or the context) for a correct decision. 

We compute this metric as the average \% drop in the model's confidence for a particular class in an image when having only the explanation map\footnote{If the model predicted the (correct) class label with the full image as input with confidence 0.8 and the model's confidence in the class fell to 0.4 when only shown the explanation map, the \% drop in model confidence would be 50\%.}. The Average Drop \% is expressed as $(\sum_{i=1}^N\frac{max(0,Y^c_i - O^{c}_i)}{Y^c_i})100$, where $Y^c_i$ is the model's output score (confidence) for class $c$ on the $i^{th}$ image and $O^c_i$ is the same model's confidence in class $c$ with only the explanation map region as input. We use $max$ in the numerator to handle cases where $O^c_i > Y^c_i$. This value is computed per image and averaged over the entire dataset.\\

\noindent \textbf {(ii) \% Increase in Confidence: } Complementary to the previous metric, it would be expected that there must be scenarios where providing only the explanation map region as input (instead of the full image) rather increases the confidence in the prediction (especially when the context is distracting). In this metric, we  measure the number of times in the entire dataset, the model's confidence increased when providing only the explanation map regions as input. Formally, the \% Increase in Confidence metric is defined as $(\sum_{i=1}^N\frac{\mathds{1}_{Y^c_i < O^{c}_i}}{N})100$, where $\mathds{1}_x$ is an indicator function that returns $1$ when the argument is true. All other notations are as defined for the previous metric.\\

\noindent \textbf {(iii) Win \%:} To further complement the above metrics, we also compute a metric that measures the number of times in th given set of images, the fall in the model's confidence for an explanation map generated by Grad-CAM++ is less (more favorable) than that of Grad-CAM. This value is expressed as a percentage.\\

The results of our experiments on the ImageNet (ILSVRC2012) validation dataset are shown in Table \ref{ImageNet results}. Grad-CAM++ performs better than Grad-CAM on all three metrics. A higher \% increase in confidence and a lower average drop \% is consistent with our hypothesis that the pixel-wise weighting adopted by Grad-CAM++ in generating the visual explanations is more \textit{model-appropriate} and consistent with the model's prediction. We also performed the same experiment on the Pascal VOC 2007 validation set. 
The results for this experiment are shown in Table \ref{VOC results}, which once again supports the superior performance of Grad-CAM++. In this case, the Pascal VOC 2007 train set was used to fine-tune the VGG-16 network (trained on ImageNet). 
\begin{table}
\begin{center}
\begin{tabular}{|l|c|c|}
\hline
Method & \textbf{Grad-CAM++} & Grad-CAM   \\
\hline\hline
Average Drop \% & \textbf{36.84} & 46.56\\
\textit{(Lower is better)} &  & \\
\hline
\% Incr. in Confidence & \textbf{17.05} & 13.42 \\
\textit{(Higher is better)} &  & \\
\hline
Win \% & \textbf{70.72} & 29.28\\
\textit{(Higher is better)} &  & \\
\hline
\end{tabular}
\end{center}
\caption{Results for objective evaluation of the explanations generated by Grad-CAM++ and Grad-CAM on the ImageNet (ILSVRC2012) validation set (``incr''=increase).}
\label{ImageNet results}
\vspace{-6mm}\end{table}
\begin{table}
\begin{center}
\begin{tabular}{|l|c|c|}
\hline
Method & \textbf{Grad-CAM++} & Grad-CAM   \\
\hline\hline
Average Drop \% & \textbf{19.53} & 28.54\\
\textit{(Lower is better)} &  & \\
\hline
\% Incr. in Confidence & 18.96 & \textbf{21.43} \\
\textit{(Higher is better)} &  & \\
\hline
Win \% & \textbf{61.47} & 39.44\\
\textit{(Higher is better)} &  & \\
\hline
\end{tabular}
\end{center}
\caption{Results for objective evaluation of the explanations generated by Grad-CAM++ and Grad-CAM on the PASCAL VOC 2007 validation set (``incr''=increase).}
\label{VOC results}
\vspace{-6mm}\end{table}

More empirical results showing the effectiveness of Grad-CAM++ for other architectures, viz, AlexNet \cite{krizhevsky2012imagenet} and Resnet-50 \cite{he2016deep} are provided in Appendices \ref{Supplementary:B} and \ref{Supplementary:C}.

\subsection{Evaluating Human Trust}
\label{Section:Evaluating Trust}
\begin{figure*}[t]
\begin{centering}
   \includegraphics[scale=2.3]{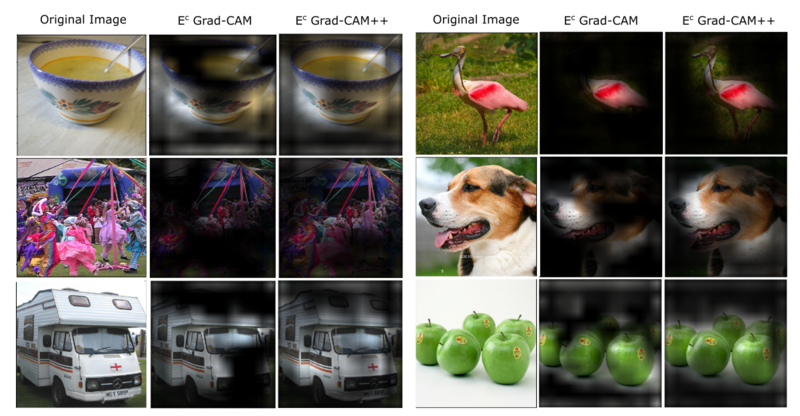}
   \caption{Sample visual explanations on ImageNet generated by Grad-CAM and Grad-CAM++ \textit{(Section \ref{Section:Evaluating Trust})}}
\label{fig:more explanation map}
\end{centering}
\vspace{-5mm}\end{figure*}
In the previous subsection, we explored the \textit{faithfulness} of the proposed method; here, we evaluate the human interpretability or \textit{trust} of our explanations. We generated explanation maps for all images in the ImageNet validation set for 5 classes, leading to a total of 250 images. The explanations generated by Grad-CAM were treated as baseline for comparison. These maps, along with their corresponding original image, were shown to 13 human subjects (who have no knowledge of the field or deep learning whatsoever) and asked which explanation algorithm invoked more trust in the underlying model. The explanation algorithm that gets more votes from the subjects can be considered as invoking more human trust in the underlying VGG-16 model. 
To further substantiate our claim, we chose 5 classes which have the highest F1-score for the validation dataset (above 0.94). As each class just has 50 images in the validation set, F1-score (harmonic mean of precision and recall) is a better suited metric than classification error. 

For each image, two explanation maps were generated, one from Grad-CAM and one from Grad-CAM++. Examples of some of these explanation maps are presented in Fig \ref{fig:more explanation map}. The subjects were provided the class of the image and asked to select the map they felt best described the object in the image (without knowledge of which one is Grad-CAM or Grad-CAM++). The subjects also had the option to select ``same" if they felt both the generated explanation maps were similar. The responses for each image was normalized, such that the total score possible for each image is 1.0\footnote{To elaborate on this point, we obtained 13 responses for each image. For example, among the 13 responses, if 5 chose the explanation map generated by Grad-CAM++, 4 chose the explanation map generated by Grad-CAM and 4 chose the option ``same", the respective scores from Grad-CAM++ and Grad-CAM would be 0.38 and 0.31 (with the remaining being ``same'').}. These normalized scores were then added, with the total achievable score being 250. Grad-CAM++ achieved a score of \textbf{109.69} as compared to \textbf{56.08} of Grad-CAM. The remaining 84.23 was labeled as "same" by the subjects. This empirical study provides strong evidence for our hypothesis that the proposed improvement in Grad-CAM++ helps aid human-interpretable image localization, and thus invokes greater trust in the model that makes the decision. As Grad-CAM++ is a generalization of Grad-CAM, it performs similar to Grad-CAM in about 33.69\% cases. 

\subsection{Harnessing Explanations for Object Localization}
\label{Section:Object localization}
\begin{figure*}[h]
\begin{center}
   \includegraphics[scale=1.15]{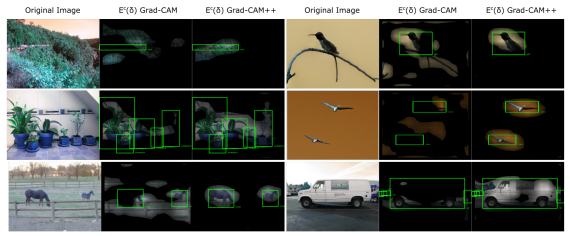}
\end{center}
   \caption{Object localization capabilities of Grad-CAM and Grad-CAM++, shown for $E^c(\delta = 0.25)$. The green boxes represent ground truth annotations for the images. \textit{(Section \ref{Section:Object localization})}}
\label{fig:object localization map}
\vspace{-5mm}\end{figure*}
In this subsection, we show the effectiveness of Grad-CAM++ for class-conditional localization of objects in a given image. We selected Pascal VOC 2012 dataset for this experiment as it has bounding box annotations for each of its image. The VGG-16 network was fine-tuned on the VOC 2012 train set and evaluations were performed on the VOC 2012 validation set. For a given image and a class $c$, the corresponding explanation map $E^c(\delta)$ is generated using Eqn \ref{explaination_maps}, with a slight modification that the class-specific saliency maps $L^c$s are min-max normalized and thresholded by an intensity $\delta$, i.e. all intensities above $\delta$ were converted to 1.0. We define an Intersection over Union (IoU) metric $Loc^c_I(\delta)$, for a class $c$, threshold value $\delta$ and an image $I$, as: 
\begin{equation}
 Loc^c_I(\delta) = \frac{Area (internal \ pixels)}{Area(bounding \ box) + Area(external \ pixels)} 
\end{equation}
where $Area (bounding \ box)$ refers to the area of the bounding box/es for a class $c$ in a given image $I$, $Area (internal \ pixels)$ refers to the number of non-zero pixels in the explanation map that lie inside the bounding box/es and $Area (external \ pixels)$ refers to the number of non-zero pixels that lie outside the bounding box/es. Higher the value of $Loc^c_I(\delta)$, better the localization of the explanation map. We only considered those images in the VOC 2012 val set, which contained bounding box annotations for at least one class in the predicted Top-5 classes by the VGG-16 network. The results for this experiment are presented in Table \ref{Localization results}. The same $\delta$ was used to threshold both explanation maps (Grad-CAM++ and Grad-CAM) for fairness of comparison. The results show Grad-CAM++'s improvement over Grad-CAM on this metric too. In particular, the IoU improvement increases with greater values of $\delta$, which supports our intuition presented in Section \ref{Section:Intuition} that Grad-CAM's heatmaps of the objects have lower intensity values in general, when compared to Grad-CAM++. Examples of the improved object localization obtained by Grad-CAM++ are shown in Fig \ref{fig:object localization map}. 
\begin{table}
\begin{center}
\begin{tabular}{|l|c|c|}
\hline
Method & \textbf{Grad-CAM++} & Grad-CAM   \\
\hline\hline
m$Loc^c_I (\delta = 0)$ & \textbf{0.34} & 0.33\\
\hline
m$Loc^c_I (\delta = 0.25)$ & \textbf{0.38} & 0.28\\
\hline
m$Loc^c_I (\delta = 0.5)$ & \textbf{0.28} & 0.16\\
\hline
\end{tabular}
\end{center}
\caption{IoU results for object localization on the PASCAL VOC 2012 val set (higher is better). m$Loc^c_I(\delta = \eta)$ refers to mean $Loc^c_I(\delta)$ per label per image, with a threshold value of $\delta = \eta$. \textit{(Section \ref{Section:Object localization})}}
\label{Localization results}
\vspace{-6mm}\end{table}

\section{Learning from Explanations: Knowledge Distillation}
\label{Section: teacher-student}
Inspired by the ideas introduced by Zagoruyko and Komodakis \cite{zagoruyko2016paying}, we show that in a constrained teacher-student learning setting \cite{hinton2015distilling,romero2014fitnets,sau2016deep,you2017learning}, knowledge transfer to a shallow student (commonly called \textit{knowledge distillation}) is possible from the explanation of CNN decisions generated by Grad-CAM++. For the first experiment, we use Wide Resnets \cite{zagoruyko2016wide} for both the student and teacher networks. We train a WRN-40-2 teacher network (2.2 M parameters) on the CIFAR-10 \cite{krizhevsky2014cifar} dataset. In order to train a student WRN-16-2 network (0.7 M parameters), we introduce a modified loss $L_{exp\_student}$, which is a weighted combination of the standard cross entropy loss $L_{cross\_ent}$ and an \textit{interpretability loss}, $L_{interpret}$: 
\begin{multline}
	L_{exp\_student}(c, W_{s}, W_{t}, I) = L_{cross\_ent}(c, W_{s}(I)) + \\ 
\alpha(L_{interpret}(c, W_{s} ,W_{t}, I))
    \label{student_loss}
\end{multline}
where $L_{interpret}$ is defined as:
\begin{equation}
	L_{interpret}(c, W_{s} ,W_{t}, I) = ||L^c_{s}(W_s(I)) - L^c_{t}(W_t(I))||^2_2
    \label{interpretability_loss}
\end{equation}
\noindent In the above equations, $I$ refers to the input image and $c$ denotes the corresponding output class label.  $L^c$ is as defined in Eqn \ref{saliency_maps} and $\alpha$ is a hyper parameter that controls the importance given to the interpretability loss. $W_s$ refers to the weights of the student network, and $W_t$ the weights of the teacher network. The intuition behind our formulation in Eqn \ref{student_loss} is that the student network should not only minimize standard cross-entropy loss for classification, but also learn from the most relevant parts of a given image used for making a decision (the $L_{interpret}$ term) from the teacher network.

Table \ref{teacher student results} shows the results for this experiment. $L_{exp\_student}$(Grad-CAM++) and $L_{exp\_student}$(Grad-CAM) refer to loss functions as defined in Eqn \ref{student_loss}, where the explanations for image $I$ are generated using Grad-CAM++ and Grad-CAM respectively. $L_{cross\_ent}$ is the normal cross entropy loss function, i.e. the student network is trained independently on the dataset without any intervention from the expert teacher. The first three rows show these results. We further also included $L_{KD}$, the knowledge distillation loss introduced by Hinton \etal in \cite{hinton2015distilling} with temperature parameter set to 4 (same as used in \cite{zagoruyko2016paying}), and these results are shown in Rows 4-6 of Table \ref{teacher student results}. The original teacher's error rate was $5.8\%$. These results show that: (i) knowledge distillation can be improved by considering the explanations of the teacher; and (ii) Grad-CAM++ provides better explanation-based knowledge distillation than Grad-CAM. We note that the student considered had a $68.18\%$ reduction in the number of parameters when compared to the teacher in this experiment.

To further study the potential of knowledge distillation using Grad-CAM++, we conducted experiments on the PASCAL VOC 2007 data set, and the results are shown in Table \ref{teacher student VOC}. $L_{cross\_ent}$ is once again the normal cross entropy loss function, i.e. the student network is trained independently on the dataset without any intervention from the expert teacher.  In the CIFAR-10 dataset, each image is of size $32  \times  32$, allowing little spatial bandwidth for transfer of salient explanations \cite{zagoruyko2016paying}. However, the VOC 2007 data set has larger images with bigger spatial extents of the visual explanations for a CNN's decision. The results show an increase in the mean Average Precision (mAP) of about $35\%$ as compared to training the student network solely on the VOC 2007 train set. The teacher network is a standard VGG-16 architecture pretrained on Imagenet with the penultimate layer fine-tuned to the VOC 2007 train set. The student network was a shallower 11-layer CNN with 27M parameters (an $80\%$ reduction). (The $\alpha$ parameter in Eqn \ref{student_loss} was taken to be 0.01 for all experiments in this section). 

\begin{table}
\begin{center}
\begin{tabular}{|l|c|}
\hline
Loss function used & Test error rate \\
\hline\hline
$L_{cross\_ent}$ & 6.78 \\
\hline
\textbf{$L_{exp\_student}$(Grad-CAM++)} & \textbf{6.74} \\
\hline
$L_{exp\_student}$(Grad-CAM) & 6.86 \\
\hline \hline
$L_{cross\_ent} + L_{KD}$ & 5.68 \\
\hline
\textbf{$L_{exp\_student}$(Grad-CAM++)$ + L_{KD}$} & \textbf{5.56} \\
\hline
$L_{exp\_student}$(Grad-CAM)$ + L_{KD}$ & 5.8 \\
\hline
\end{tabular}
\end{center}
\caption{Results for knowledge distillation to train a student (WRN-16-2) from a deeper teacher network (WRN-40-2). (Section \ref{Section: teacher-student} contains the description of the loss functions used.) }
\label{teacher student results}
\vspace{-6mm}\end{table}

\begin{table}
\begin{center}
\begin{tabular}{|l|c|}
\hline
loss function used & mAP (\% increase) \\
\hline\hline
\textbf{$L_{exp\_student}$(Grad-CAM++)} & \textbf{0.42 (35.5\%)} \\
\hline
$L_{cross\_ent} + L_{KD}$ & 0.34 (9.7\%) \\
\hline
$L_{cross\_ent}$ [Baseline] & 0.31 (0.0\%) \\
\hline
\end{tabular}
\end{center}
\caption{Results for training a student network with explanations from the teacher (VGG-16 fine-tuned) and with knowledge distillation on PASCAL VOC 2007 dataset. The \% increase is with respect to the baseline loss $L_{cross\_ent}$. \textit{(Section \ref{Section: teacher-student})}}
\label{teacher student VOC}
\vspace{-6mm}\end{table}

\section{Explanations for Image Captioning and 3D Action Recognition}
\label{Section: Other visual tasks}
Similar to other such methods, Grad-CAM++ can be used to understand any machine learning model's decision as long as it utilizes a CNN as an integral module. In this section, we present the results for experiments on two such tasks - Image Captioning and 3D Action Recognition. To the best of our knowledge, this is the first effort to generate visual explanations of CNNs in the video domain.

\subsection{Image Captioning}
\label{Section:Image Captioning}
\begin{figure*}[t]
\begin{center}
   \includegraphics[scale=0.4]{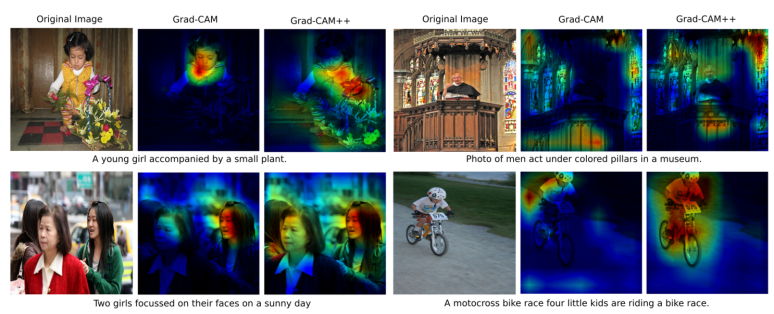}
\end{center}
   \caption{Visual explanations of image captions predicted by CNN-based neural network architectures using both Grad-CAM and Grad-CAM++. \textit{(Section \ref{Section:Image Captioning})}}
\label{fig:caption map}
\vspace{-5mm}\end{figure*}
We considered a standard image captioning model~\cite{Puri2017} trained on the Flickr30k data set~\cite{Plummer2017}~\cite{Young2014} using an adaptation of the popular Show-and-Tell model~\cite{Vinyals2015}. The architecture includes a CNN to encode the image followed by an LSTM to generate the captions. For fairness of comparison, we use the same settings as used for this task in Grad-CAM. To generate the heatmaps, Eqn \ref{saliency_maps} was used with the log probability of the predicted caption as $Y^c$ (for classification tasks, $Y^c$ was related to the output of the neuron representing the $c^{th}$ class). Fig \ref{fig:caption map} illustrates the visual explanations of four randomly chosen images from the Flickr30k data set for the predicted caption. In all the images, Grad-CAM++ produces more complete heatmaps than Grad-CAM.
For instance, in the first example, Grad-CAM++ highlights both the girl and the plant for the caption ``A young girl accompanied by a small plant", whereas Grad-CAM highlights only the girl. In the second example in the first row, although the predicted caption is wrong, Grad-CAM++'s visualization gives insight into what the network focused on - the colored glasses (which is predicted as pillars by the network) and the man. Comparatively, Grad-CAM's visualization is incomplete with no heat generated at the man's spatial location in the image.
\begin{figure*}[t]
\begin{center}
   \includegraphics[scale=1.2]{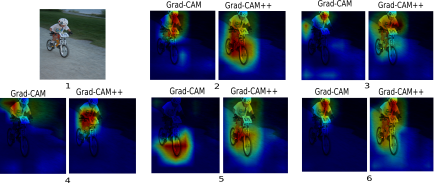}
\end{center}
   \caption{Visual explanations generated by Grad-CAM and Grad-CAM++ on a given image (1) predicting 5 different captions: (2) A little boy rides a bike down a hill on a miniature dirt bike; (3) A young boy in a helmet rides a bike on the road; (4) A child with a helmet on his head rides a bike; (5) The young boy pedals quickly at a BMX race; and (6) The little boy rides his bicycle in a race.}
\label{fig:caption result 1}
\vspace{-5mm}\end{figure*}
In order to study the experiment further with diverse captions, we generated visual explanations for the four images using 5 different captions, which were used while training the captioning model~\cite{Puri2017}. We show one such result in Figure \ref{fig:caption result 2}. The results with the other three images are shown in Appendix Figures \ref{fig:caption result 1}, \ref{fig:caption result 3} and \ref{fig:caption result 4}. For all the captions, Grad-CAM++ provides more complete explanations corresponding to the captions when compared to Grad-CAM.

\subsection{3D Action Recognition}
\label{Section:3D Action Recognition}
\begin{figure*}[t]
\begin{center}
   \includegraphics[scale=1.5]{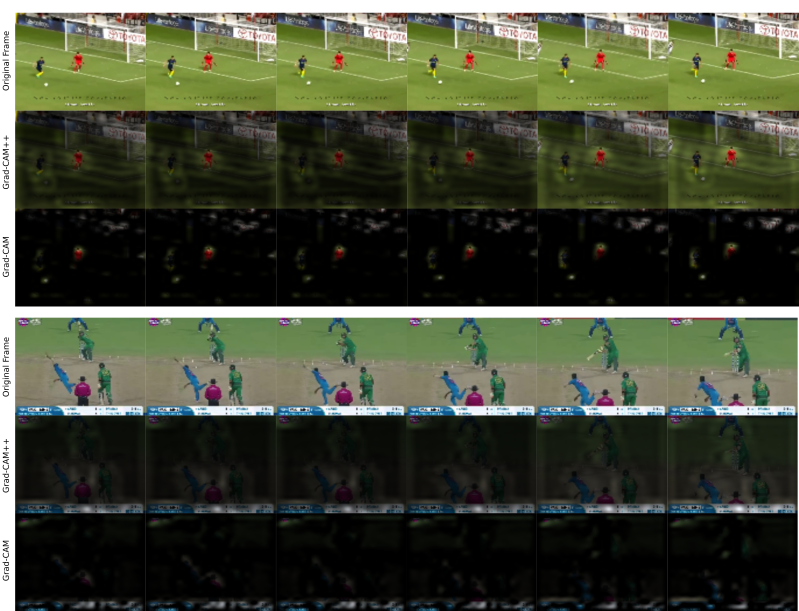}
\end{center}
   \caption{Example explanation maps $E^c$ (see Eqn \ref{explaination_maps}) for video frames generated by Grad-CAM and Grad-CAM++ for a particular predicted action. The first set of video frames correspond to the predicted action "indoor soccer" and the next set of video frames correspond to "one day international". In the "indoor soccer" example, the $E^c$s generated by Grad-CAM++ highlight the entire football ground with special emphasis on the two players, while Grad-CAM only highlights parts of the player's body. In the second set of video frames, the Grad-CAM++ explanations highlight the scoreboard with some importance to the players and pitch. The Grad-CAM interpretation of the model's decision on the other hand is bland.}
\label{fig:video examples}
\vspace{-5mm}\end{figure*}

\begin{table}
\begin{center}
\begin{tabular}{|l|c|c|}
\hline
Method & \textbf{Grad-CAM++} & Grad-CAM   \\
\hline\hline
Average Drop \% & \textbf{59.79} & 95.26\\
\textit{(Lower is better)} &  & \\
\hline
\% Incr. in Confidence & \textbf{6.68} & 0.84\\
\textit{(Higher is better)} &  & \\
\hline
Win \% & \textbf{94.09} & 5.91\\
\textit{(Higher is better)} &  & \\
\hline
\end{tabular}
\end{center}
\caption{Results on the 3D action recognition task for visual explanations  generated by Grad-CAM++ and Grad-CAM for 3k videos from the Sports-1M Dataset using same performance metrics introduced in Section \ref{Section:Objective evaluation of performance based on object recognition} (``incr'' denotes increase).}
\label{Video results}
\vspace{-6mm}\end{table}
For the task of 3D action recognition, we used a 3D-CNN architecture, in particular, the C3D model \cite{tran2015learning} trained on the Sports-1M Dataset \cite{KarpathyCVPR14}, which contains $1,133,158$ YouTube sports videos annotated with $487$ Sports labels. In particular, we selected windows of 16 frames from each video to train the model. The visual explanations are generated from the last convolution layer feature maps of the C3D model. The generated explanations (for both Grad-CAM and Grad-CAM++) were upsampled to the video resolution and then the corresponding explanation video maps were generated by point-wise multiplication with the original video. While generating the explanation maps $E^c$ as in Eqn \ref{explaination_maps}, $I$ is the input video and $c$ is the predicted action of the video. For empirical evaluation of the generated explanations, we collected arbitrary 3k videos from Sports-1M Datsaet \cite{KarpathyCVPR14} and used the same performance metrics described in Section \ref{Section:Empirical evaluation of generated explanations}. The results of our experiment on these randomly selected 3k videos are shown in Table \ref{Video results}. The performance of Grad-CAM++ is better than Grad-CAM in all the metrics, thus supporting Grad-CAM++'s merit for use in video-based classification tasks. Sample qualitative results are shown in Figure \ref{fig:video examples}, where the 16 frames in a video is subsampled to 6 frames for clearer presentation (no handpicking was done while choosing these videos). In both scenarios, the explanations generated by Grad-CAM++ are more semantically relevant. (We also enclose MPEG video files as supplementary materials, which provide a better visualization of our claim. The file name indicates the action predicted by the CNN model as well as the method used. For example, \texttt{tennis.mpg}, \texttt{tennis\_gcam.mpg} and \texttt{tennis\_gcam++.mpg} refer to the original 16 frame video, the explanation maps generated using Grad-CAM and explanation maps generated using Grad-CAM++ respectively when the model predicted ``tennis'' as the action.)
In general, the generated video explanation maps, $E^c$s, show a clearer explanation for the corresponding action in the case of Grad-CAM++ as compared to Grad-CAM. Grad-CAM++ tends to highlight the context of the video (similar to images) as less bright and most discriminative parts as brighter regions in the video explanations. The quantitative results in Table \ref{Video results} suggest that the region highlighted by Grad-CAM++ is more relevant for the prediction made by the model. 

\section{Discussion}
\label{sec_discussion}
\begin{figure}[h]
\begin{center}
   \includegraphics[scale=0.45]{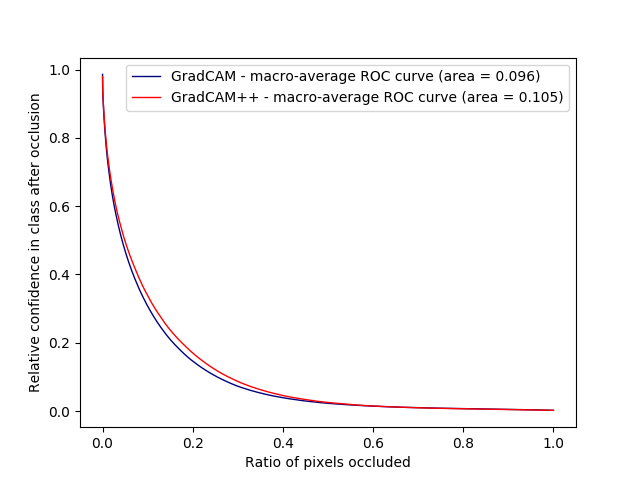}
\end{center}
   \caption{ROC curve to study the relationship between spatial extents of visual explanations and the corresponding relative confidence when the visual explanation region is provided as input to the model. \textit{(See Section \ref{subsec_larger_maps_discussion} for more details.)}}
\label{fig:ROC_MACRO}
\vspace{-5mm}\end{figure}
\subsection{Why only Positive Gradients in Grad-CAM++?}
\label{subsec_why_positive_grads}
In Section \ref{Section:Approach}, we hypothesize that a weighted combination of positive gradients w.r.t. each pixel in an activation map $A^k$ strongly correlates with the importance of that activation map for a given class $c$ (Eqn \ref{weighted version}). In this section, we test the correctness of this assumption by relaxing the constraint on the gradients. We take the same VGG-16 model used for our objective evaluation studies on the Pascal VOC 2007 val set (Section \ref{Section:Objective evaluation of performance based on object recognition}) and redo the experiment with a slightly different $w^c_k$:  
\begin{equation}
 w_k^c = \sum_{i}\sum_{j}\alpha_{ij}^{kc}.\frac{\partial Y^c}{\partial A_{ij}^k}\label{weighted version all gradients}
\end{equation}
Note that the \texttt{relu()} function is dropped as we consider all the gradients. The $\alpha_{ij}^{kc}$s are calculated as in Eqn \ref{eq_final_alpha}, with the exception that $\alpha_{ij}^{kc} \neq 0$ for negative gradients.
\begin{equation}
 \alpha_{ij}^{kc} = \frac{\frac{\partial ^2Y^c}{(\partial A_{ij}^k)^2}}{2 \frac{\partial ^2Y^c}{(\partial A_{ij}^k)^2}\! + \! \sum_{a}\sum_{b} A_{ab}^k\{\frac{\partial ^3Y^c}{(\partial A_{ij}^k)^3}\}}
\end{equation}
We refer to this modified version of Grad-CAM++ (where we do not restrict to positive gradients) as Grad-CAM++$^{\perp}$. Table \ref{supplementary all gradients} shows the poor performance of Grad-CAM++$^{\perp}$ when compared to Grad-CAM. These results support our claim that the positive gradients are critical to decide the importance of an activation map $A^k$ for a given class $c$.

\begin{table}[!htbp]
\begin{center}
\begin{tabular}{|l|c|c|}
\hline
Method & Grad-CAM++$^{\perp}$ & \textbf{Grad-CAM}  \\
\hline\hline
- Average drop\% & 32.43 & \textbf{28.54}\\
\textit{(Lower is better)} &  & \\
\hline
- \% incr. in confidence & 19.12 & \textbf{21.43} \\
\textit{(Higher is better)} &  & \\
\hline
- Win\% & 26.09 & \textbf{73.91}\\
\textit{(Higher is better)} &  & \\
\hline
\end{tabular}
\end{center}
\caption{Results for objective evaluation of explanations generated by Grad-CAM++$^{\perp}$ and Grad-CAM on the Pascal VOC 2007 validation set (2510 images) (``incr'' denotes increase). In this experiment, the weights $w^c_k$s were taken to be a weighed combination of all the gradients of an activation map $A^k$  (both positive and negative).}
\label{supplementary all gradients}
\vspace{-6mm}\end{table}

\subsection{Does Grad-CAM++ do well because of larger maps?}
\label{subsec_larger_maps_discussion}
One could question if Grad-CAM++ was doing well because of larger explanations in each image. In general, we expect a lower drop in classification score if the explanation map region provided as input to the model for a given image $I$ and class $c$ has greater area. We plotted an ROC curve to measure this trade-off between the spatial area of the occluded map and the relative confidence in class after occlusion (that is, the quantity $\frac{O^c_I*100}{Y^c_I}$, where $O^c_I$ is new score with occluded image and $Y^c_I$ is original score with full image as input), for both Grad-CAM and Grad-CAM++. A threshold parameter $\theta$ was varied from 0 to 1 at equally-spaced discrete intervals to generate the curve. For a given $\theta$, the occluded image is $O^c_i = I \circ \Delta$, where $\Delta_{kj} = 0$ if $L^c_{kj} < \gamma$ else  $\Delta_{kj} = 1$. $k, j$ are iterators over the pixels and $\gamma$ is the $\theta$-quantile of the empirical distribution of each explanation region's pixel values. Formally, $\Pr(L^c_{kj} < \gamma) = \theta$. The empirical distribution is calculated for each image individually and then averaged across the dataset. Figure \ref{fig:ROC_MACRO} shows the result. One can observe that at each quantile ($\theta$), Grad-CAM++ highlights regions that are as faithful or more to the underlying model than Grad-CAM, irrespective of the spatial extents. 


\section{Conclusion}
\label{Section:Conclusion}
In this work, we proposed a generalized approach for visual explanations of CNN based architectures, Grad-CAM++. We provided a derivation for our method and showed that it is a simple, yet effective generalization of earlier, popularly used, gradient-based visual explanation methods.Our method addresses the shortcomings of Grad-CAM - especially multiple occurrences of a class in an image and poor object localizations. We validated the effectiveness of our method both objectively (faithfulness to the model being explained) and subjectively (invoking human trust) using standard well-known CNN models and datasets (ImageNet and Pascal VOC). We showed that Grad-CAM++ can also prove superior on tasks such as image caption generation and video understanding (action recognition). In Section \ref{Section: teacher-student},  we motivated a research direction where explanations of a deep network are not only used to understand the reasonings behind model decisions but also utilized to train a shallower student network. The student network learned much better representations than the original teacher network (lower test error rate) when using explanations for knowledge distillation. Future work involves refining the loss formulation in the teacher-student setting so as to distill knowledge via Grad-CAM++ explanations more effectively. We also hope to study the proposed method in more detail when there are multiple classes in a single image, as well as explore the possibility of extending our algorithm to explain decisions made by other neural network architectures such as recurrent neural networks, long short-term memory networks, and generative adversarial networks.

\ifCLASSOPTIONcompsoc
  \section*{Acknowledgments}
\else
  \section*{Acknowledgment}
\fi
We thank the anonymous reviewers for their valuable feedback, which greatly helped improve the paper. We also thank the Ministry of Human Resource Development, India for financial assistance and NVIDIA for donation of K40 GPU through their Academic Hardware Grant program.


%

\bibliographystyle{IEEEtran}
\bibliography{bare_adv}

\begin{thebibliography}{10}
\providecommand{\url}[1]{#1}
\csname url@samestyle\endcsname
\providecommand{\newblock}{\relax}
\providecommand{\bibinfo}[2]{#2}
\providecommand{\BIBentrySTDinterwordspacing}{\spaceskip=0pt\relax}
\providecommand{\BIBentryALTinterwordstretchfactor}{4}
\providecommand{\BIBentryALTinterwordspacing}{\spaceskip=\fontdimen2\font plus
\BIBentryALTinterwordstretchfactor\fontdimen3\font minus
  \fontdimen4\font\relax}
\providecommand{\BIBforeignlanguage}[2]{{%
\expandafter\ifx\csname l@#1\endcsname\relax
\typeout{** WARNING: IEEEtran.bst: No hyphenation pattern has been}%
\typeout{** loaded for the language `#1'. Using the pattern for}%
\typeout{** the default language instead.}%
\else
\language=\csname l@#1\endcsname
\fi
#2}}
\providecommand{\BIBdecl}{\relax}
\BIBdecl

\bibitem{Zeiler}
M.~D. Zeiler and R.~Fergus, ``Visualizing and understanding convolutional
  networks,'' in \emph{European conference on computer vision}.\hskip 1em plus
  0.5em minus 0.4em\relax Springer, 2014, pp. 818--833.

\bibitem{Zhou}
B.~Zhou, A.~Khosla, A.~Lapedriza, A.~Oliva, and A.~Torralba, ``Learning deep
  features for discriminative localization,'' in \emph{Proceedings of the IEEE
  Conference on Computer Vision and Pattern Recognition}, 2016, pp. 2921--2929.

\bibitem{Lin}
M.~Lin, Q.~Chen, and S.~Yan, ``Network in network,'' \emph{arXiv preprint
  arXiv:1312.4400}, 2013.

\bibitem{Oquab}
M.~Oquab, L.~Bottou, I.~Laptev, and J.~Sivic, ``Is object localization for
  free?-weakly-supervised learning with convolutional neural networks,'' in
  \emph{Proceedings of the IEEE Conference on Computer Vision and Pattern
  Recognition}, 2015, pp. 685--694.

\bibitem{Pinheiro}
P.~O. Pinheiro and R.~Collobert, ``From image-level to pixel-level labeling
  with convolutional networks,'' in \emph{Proceedings of the IEEE Conference on
  Computer Vision and Pattern Recognition}, 2015, pp. 1713--1721.

\bibitem{Selvaraju}
R.~R. Selvaraju, A.~Das, R.~Vedantam, M.~Cogswell, D.~Parikh, and D.~Batra,
  ``Grad-cam: Why did you say that? visual explanations from deep networks via
  gradient-based localization,'' \emph{arXiv preprint arXiv:1610.02391}, 2016.

\bibitem{Springenberg}
J.~T. Springenberg, A.~Dosovitskiy, T.~Brox, and M.~Riedmiller, ``Striving for
  simplicity: The all convolutional net,'' \emph{arXiv preprint
  arXiv:1412.6806}, 2014.

\bibitem{bargal2018excitation}
S.~A. Bargal, A.~Zunino, D.~Kim, J.~Zhang, V.~Murino, and S.~Sclaroff,
  ``Excitation backprop for rnns,'' in \emph{Proceedings of the IEEE Conference
  on Computer Vision and Pattern Recognition}, 2018.

\bibitem{Mahendran}
A.~Mahendran and A.~Vedaldi, ``Salient deconvolutional networks,'' in
  \emph{European Conference on Computer Vision}.\hskip 1em plus 0.5em minus
  0.4em\relax Springer, 2016, pp. 120--135.

\bibitem{jason}
\BIBentryALTinterwordspacing
J.~Yosinski, J.~Clune, A.~M. Nguyen, T.~J. Fuchs, and H.~Lipson,
  ``Understanding neural networks through deep visualization,'' \emph{CoRR},
  vol. abs/1506.06579, 2015. [Online]. Available:
  \url{http://arxiv.org/abs/1506.06579}
\BIBentrySTDinterwordspacing

\bibitem{Simonyan}
K.~Simonyan, A.~Vedaldi, and A.~Zisserman, ``Deep inside convolutional
  networks: Visualising image classification models and saliency maps,''
  \emph{arXiv preprint arXiv:1312.6034}, 2013.

\bibitem{Ribeiro}
M.~T. Ribeiro, S.~Singh, and C.~Guestrin, ``Why should i trust you?: Explaining
  the predictions of any classifier,'' in \emph{Proceedings of the 22nd ACM
  SIGKDD International Conference on Knowledge Discovery and Data
  Mining}.\hskip 1em plus 0.5em minus 0.4em\relax ACM, 2016, pp. 1135--1144.

\bibitem{shrikumar2017learning}
A.~Shrikumar, P.~Greenside, and A.~Kundaje, ``Learning important features
  through propagating activation differences,'' \emph{arXiv preprint
  arXiv:1704.02685}, 2017.

\bibitem{Al-Shedivat}
M.~Al-Shedivat, A.~Dubey, and E.~P. Xing, ``Contextual explanation networks,''
  \emph{arXiv preprint arXiv:1705.10301}, 2017.

\bibitem{Konam}
S.~Konam, ``Vision-based navigation and deep-learning explanation for
  autonomy.'' in \emph{Master’s thesis, Robotics Institute, Carnegie Mellon
  University, Pittsburgh, PA.}, 2017.

\bibitem{Lengerich}
B.~J. Lengerich, S.~Konam, E.~P. Xing, S.~Rosenthal, and M.~Veloso, ``Visual
  explanations for convolutional neural networks via input resampling,''
  \emph{arXiv preprint arXiv:1707.09641}, 2017.

\bibitem{Kim}
J.~Kim and J.~Canny, ``Interpretable learning for self-driving cars by
  visualizing causal attention,'' \emph{arXiv preprint arXiv:1703.10631}, 2017.

\bibitem{Li2018}
K.~Li, Z.~Wu, K.-C. Peng, J.~Ernst, and Y.~Fu, ``Tell me where to look: Guided
  attention inference network,'' \emph{arXiv preprint arXiv:1802.10171}, 2018.

\bibitem{zhou2014object}
B.~Zhou, A.~Khosla, A.~Lapedriza, A.~Oliva, and A.~Torralba, ``Object detectors
  emerge in deep scene cnns,'' \emph{arXiv preprint arXiv:1412.6856}, 2014.

\bibitem{Simonyan1}
K.~Simonyan and A.~Zisserman, ``Very deep convolutional networks for
  large-scale image recognition,'' \emph{arXiv preprint arXiv:1409.1556}, 2014.

\bibitem{Jia}
Y.~Jia, E.~Shelhamer, J.~Donahue, S.~Karayev, J.~Long, R.~Girshick,
  S.~Guadarrama, and T.~Darrell, ``Caffe: Convolutional architecture for fast
  feature embedding,'' in \emph{Proceedings of the 22nd ACM international
  conference on Multimedia}.\hskip 1em plus 0.5em minus 0.4em\relax ACM, 2014,
  pp. 675--678.

\bibitem{krizhevsky2012imagenet}
A.~Krizhevsky, I.~Sutskever, and G.~E. Hinton, ``Imagenet classification with
  deep convolutional neural networks,'' in \emph{Advances in neural information
  processing systems}, 2012, pp. 1097--1105.

\bibitem{he2016deep}
K.~He, X.~Zhang, S.~Ren, and J.~Sun, ``Deep residual learning for image
  recognition,'' in \emph{Proceedings of the IEEE conference on computer vision
  and pattern recognition}, 2016, pp. 770--778.

\bibitem{zagoruyko2016paying}
S.~Zagoruyko and N.~Komodakis, ``Paying more attention to attention: Improving
  the performance of convolutional neural networks via attention transfer,''
  \emph{arXiv preprint arXiv:1612.03928}, 2016.

\bibitem{hinton2015distilling}
G.~Hinton, O.~Vinyals, and J.~Dean, ``Distilling the knowledge in a neural
  network,'' \emph{arXiv preprint arXiv:1503.02531}, 2015.

\bibitem{romero2014fitnets}
A.~Romero, N.~Ballas, S.~E. Kahou, A.~Chassang, C.~Gatta, and Y.~Bengio,
  ``Fitnets: Hints for thin deep nets,'' \emph{arXiv preprint arXiv:1412.6550},
  2014.

\bibitem{sau2016deep}
B.~B. Sau and V.~N. Balasubramanian, ``Deep model compression: Distilling
  knowledge from noisy teachers,'' \emph{arXiv preprint arXiv:1610.09650},
  2016.

\bibitem{you2017learning}
S.~You, C.~Xu, C.~Xu, and D.~Tao, ``Learning from multiple teacher networks,''
  in \emph{Proceedings of the 23rd ACM SIGKDD International Conference on
  Knowledge Discovery and Data Mining}.\hskip 1em plus 0.5em minus 0.4em\relax
  ACM, 2017, pp. 1285--1294.

\bibitem{zagoruyko2016wide}
S.~Zagoruyko and N.~Komodakis, ``Wide residual networks,'' \emph{arXiv preprint
  arXiv:1605.07146}, 2016.

\bibitem{krizhevsky2014cifar}
A.~Krizhevsky, V.~Nair, and G.~Hinton, ``The cifar-10 dataset,'' \emph{online:
  http://www. cs. toronto. edu/kriz/cifar. html}, 2014.

\bibitem{Puri2017}
R.~Puri and D.~Ricciardelli, ``Caption this, with tensorflow,'' in
  \emph{https://www.oreilly.com/learning/caption-this-with-tensorflow},
  Accessed 28 March, 2017.

\bibitem{Plummer2017}
B.~A. Plummer, L.~Wang, C.~M. Cervantes, J.~C. Caicedo, J.~Hockenmaier, and
  S.~Lazebnik, ``Flickr30k entities: Collecting region-to-phrase
  correspondences for richer image-to-sentence models,'' \emph{IJCV}, vol. 123,
  no.~1, pp. 74--93, 2017.

\bibitem{Young2014}
P.~Young, A.~Lai, M.~Hodosh, and J.~Hockenmaier, ``From image descriptions to
  visual denotations: New similarity metrics for semantic inference over event
  descriptions,'' \emph{Transactions of the Association for Computational
  Linguistics}, pp. 67--78, 2014.

\bibitem{Vinyals2015}
O.~Vinyals, A.~Toshev, S.~Bengio, and D.~Erhan, ``Show and tell: A neural image
  caption generator,'' in \emph{Proceedings of the IEEE Conference on Computer
  Vision and Pattern Recognition}.\hskip 1em plus 0.5em minus 0.4em\relax IEEE,
  2015, pp. 3156--3164.

\bibitem{tran2015learning}
D.~Tran, L.~Bourdev, R.~Fergus, L.~Torresani, and M.~Paluri, ``Learning
  spatiotemporal features with 3d convolutional networks,'' in \emph{Computer
  Vision (ICCV), 2015 IEEE International Conference on}.\hskip 1em plus 0.5em
  minus 0.4em\relax IEEE, 2015, pp. 4489--4497.

\bibitem{KarpathyCVPR14}
A.~Karpathy, G.~Toderici, S.~Shetty, T.~Leung, R.~Sukthankar, and L.~Fei-Fei,
  ``Large-scale video classification with convolutional neural networks,'' in
  \emph{Proceedings of the IEEE Conference on Computer Vision and Pattern
  Recognition}, 2014.

\end{thebibliography}


\begin{IEEEbiography}[{\includegraphics[width=1in,height=1.25in,clip,keepaspectratio]{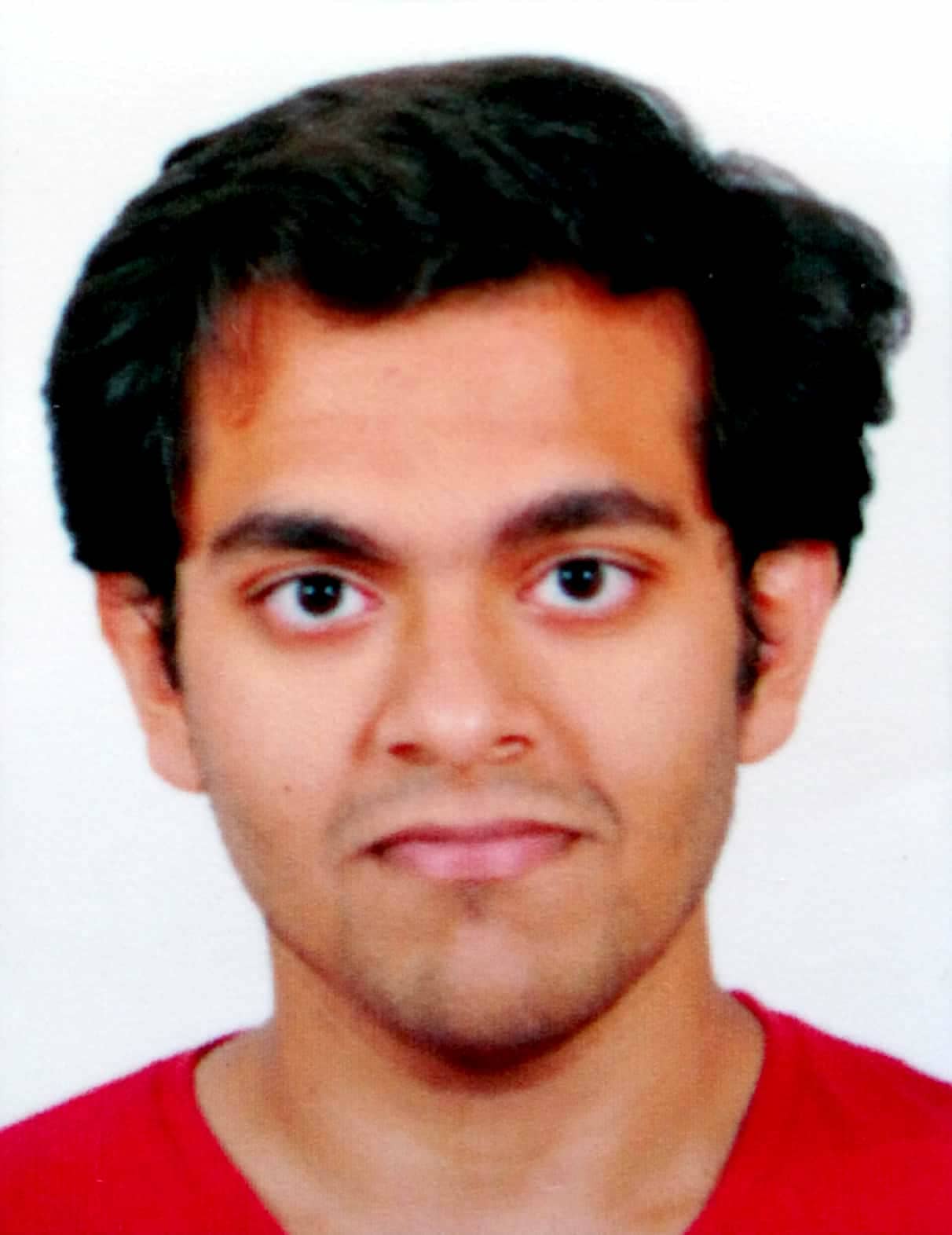}}]{Aditya Chattopadhyay is currently a Research Assistant in the Department of Computer Science and Engineering, Indian Institute of Technology Hyderabad, India. He received the Bachelor of Technology
degree in Computer Science and Master of Science by Research degree in Computational Natural Sciences from the International Institute of Information Technology, Hyderabad in 2016 and 2018 respectively. His research interests include
explainable Artificial Intelligence, statistical modeling and theoretical aspects of machine learning with a special focus on optimization. He was awarded the Gold Medal for Academic Excellence in the master`s  program  in computational natural sciences in 2018.}
\vspace{-8pt}
\end{IEEEbiography}
\begin{IEEEbiography}[{\includegraphics[width=1in,height=1.25in,clip,keepaspectratio]{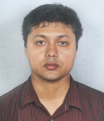}}]{Anirban Sarkar is currently a doctoral student in the Department of Computer Science and Engineering, Indian Institute of Technology Hyderabad, India. He received the Master of Technology degree in computer science from National Institute of Technology, Rourkela, India, in 2016. He worked in IBM India as systems engineer for two and half years before joining the Masters program. His research interests include machine learning for computer vision, explainability of machine learning models and applications of causality in machine learning with a specific focus on deep learning.}
\end{IEEEbiography}
\vspace{-8pt}
\begin{IEEEbiography}[{\includegraphics[width=1in,height=1.25in,clip,keepaspectratio]{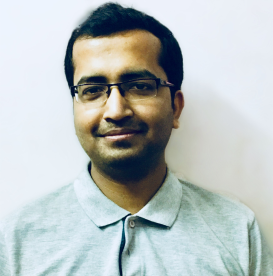}}]{Prantik Howlader is currently a software engineer and machine learning researcher in the Cisco Security Business Group, India. He received the M.Tech degree in computer science from National Institute of Technology, Calicut, India, in 2016. His research interests include Security, explainable AI and machine learning.}
\end{IEEEbiography}
\vspace{-8pt}
\begin{IEEEbiography}[{\includegraphics[width=1in,height=1.25in,clip,keepaspectratio]{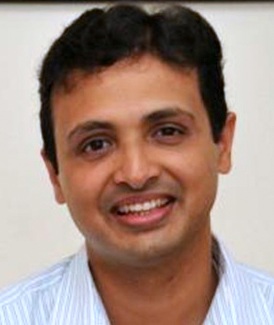}}]{Vineeth N Balasubramanian is an Associate Professor in the Department of Computer Science and Engineering at the Indian Institute of Technology, Hyderabad, India. His research interests include deep learning, machine learning, computer vision, non-convex optimization and real-world applications in these areas. He has over 60 research publications in premier peer-reviewed venues including CVPR, ICCV, KDD, ICDM, IEEE TPAMI and ACM MM, as well as an edited book on a recent development in machine learning called Conformal Prediction. His PhD dissertation at Arizona State University on the Conformal Predictions framework was nominated for the Outstanding PhD Dissertation at the Department of Computer Science. He is an active reviewer/contributor at many conferences such as ICCV, IJCAI, ACM MM and ACCV, as well as journals including IEEE TPAMI, IEEE TNNLS, Machine Learning and Pattern Recognition.}
\end{IEEEbiography}

\newpage
\clearpage
\pagenumbering{arabic}

\appendices
\section{Supplementary Materials}
We herein present further experimental results to confirm the effectiveness of Grad-CAM++.
In the main text, we carried out all experiments with the VGG-16 architecture. To show that our results are not biased by the choice of CNN architecture, we carry out extensive experiments using both AlexNet \cite{krizhevsky2012imagenet} and ResNet-50 \cite{he2016deep} architectures. For all experiments, the activation maps $A^k$ considered were the last convolutional feature maps of the network (as in the main text for VGG-16). This appendix is organized as follows:
\begin{itemize}
  \item We first present results from objective evaluation of performance of the two explanation algorithms, viz Grad-CAM and Grad-CAM++ for both AlexNet and ResNet-50 architectures, similar to the one carried out in Section \ref{Section:Objective evaluation of performance based on object recognition} of the main text. 
  \item This is followed by additional qualitative results of our Image Captioning results from Section \ref{Section:Image Captioning}.
\end{itemize}

\subsection{Evaluation of Object Recognition Performance with Other Architectures}
\label{Supplementary:B}
\begin{figure*}[t]
\begin{center}
   \includegraphics[scale=1.8]{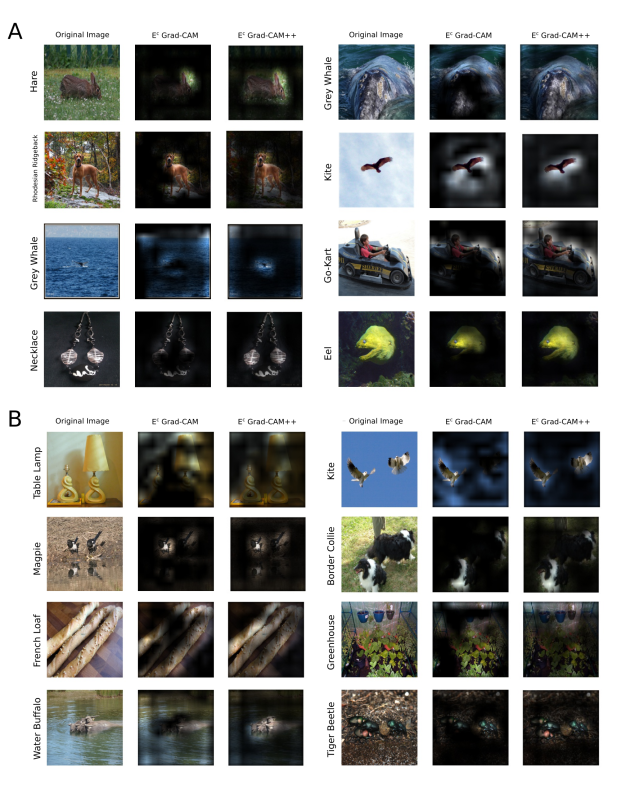}
\end{center}
   \caption{Example explanation maps $E^c$ (see Eqn \ref{explaination_maps} in main text) for images generated by Grad-CAM and Grad-CAM++. These explanations are for decisions made by the \textbf{AlexNet} architecture. Panel A shows images where Grad-CAM++ solves the problem of poor class localization. Panel B depicts images where Grad-CAM++ is effective for explaining multiple occurrences of the same class in an image. For each set of three images the class label predicted by the network is written  horizontally on the leftmost edge.}
\label{fig:alexnet explanation map}
\vspace{-5mm}\end{figure*}

\begin{figure*}[t]
\begin{center}
   \includegraphics[scale=1.8]{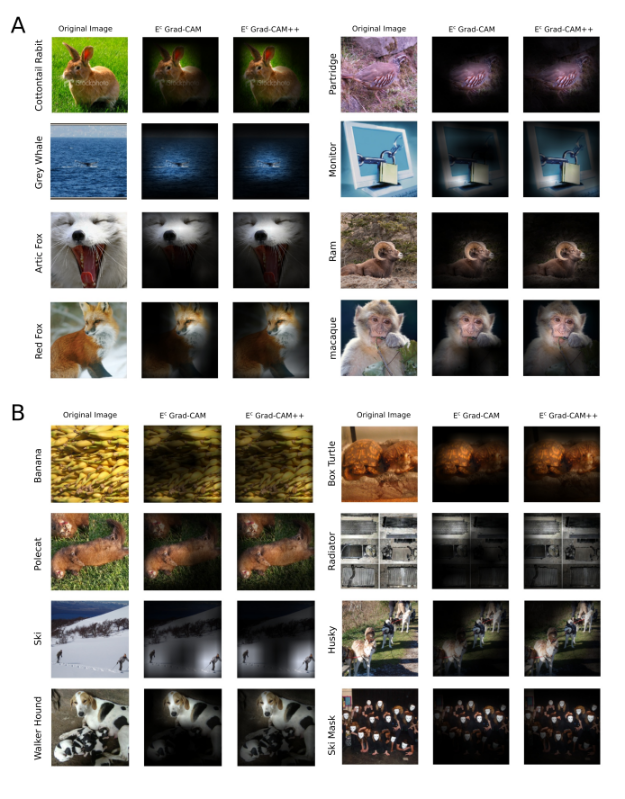}
\end{center}
   \caption{Example explanation maps $E^c$ (see Eqn \ref{explaination_maps} in main text) for images generated by Grad-CAM and Grad-CAM++. These explanations are for decisions made by the \textbf{ResNet-50} architecture. Panel A shows images where Grad-CAM++ solves the problem of poor class localization. Panel B depicts images where Grad-CAM++ is effective for explaining multiple occurrences of the same class in an image. For each set of three images the class label predicted by the network is written horizontally on the leftmost edge.}
\label{fig:resnet explanation map}
\vspace{-5mm}\end{figure*}

The experimental setup in this section is the same as described in Section \ref{Section:Objective evaluation of performance based on object recognition} of the main text, with the only difference being the CNN architecture used. Tables \ref{supplementary alexnet imagenet} and \ref{supplementary alexnet pascal} report results for experiments carried out using AlexNet, while Tables \ref{supplementary resnet imagenet} and \ref{supplementary resnet pascal} present the empirical results for the ResNet-50 architecture. These results follow a similar trend as seen in Tables \ref{ImageNet results} and \ref{VOC results} in the main text, and further accentuate our claim of the improved performance obtained by Grad-CAM++. 
Qualitative results are shown in Fig \ref{fig:alexnet explanation map} for AlexNet and in Fig \ref{fig:resnet explanation map}for the ResNet-50 architecture.

We note in passing that there seems to be a correlation between the "Average drop \%" metric (for both Grad-CAM and Grad-CAM++) and the generalization prowess of the deep network they explain. Resnet-50 has the lowest top-1 and top-5 classification error on the ImageNet dataset, followed by VGG-16 and subsequently AlexNet \cite{he2016deep,Simonyan1}. For ImageNet the "Average drop \%" metric for ResNet-50 ($28-31\%$), VGG-16 ($36-47\%$) and AlexNet ($62-83\%$) follows the same trend. This correlation also exists for the Pascal VOC dataset. \textit{This metric can hence be potentially exploited to help obtain more generalizable deep network models from data}. 

\begin{table}[h]
\begin{center}
\begin{tabular}{|l|c|c|}
\hline
Method & \textbf{Grad-CAM++} & Grad-CAM   \\
\hline\hline
- Average drop\% & \textbf{62.75} & 82.86\\
\textit{(Lower is better)} &  & \\
\hline
- \% incr. in confidence & \textbf{8.24} & 3.16 \\
\textit{(Higher is better)} &  & \\
\hline
- Win\% & \textbf{86.56} & 13.44\\
\textit{(Higher is better)} &  & \\
\hline
\end{tabular}
\end{center}
\caption{Results for objective evaluation of the explanations generated by both Grad-CAM++ and Grad-CAM on the ImageNet (ILSVRC2012) validation set (``incr'' denotes increase). The explanations were generated for decisions taken by the \textbf{AlexNet} architecture. \textit{(See Section \ref{Section:Objective evaluation of performance based on object recognition} for details of the metrics used.)}}
\label{supplementary alexnet imagenet}
\vspace{-4mm}\end{table}
\begin{table}[h]
\begin{center}
\begin{tabular}{|l|c|c|}
\hline
Method & \textbf{Grad-CAM++} & Grad-CAM   \\
\hline\hline
- Average drop\% & \textbf{29.16} & 45.82\\
\textit{(Lower is better)} &  & \\
\hline
- \% incr. in confidence & \textbf{19.76} & 14.38 \\
\textit{(Higher is better)} &  & \\
\hline
- Win\% & \textbf{72.79} & 27.21\\
\textit{(Higher is better)} &  & \\
\hline
\end{tabular}
\end{center}
\caption{Results for objective evaluation of the explanations generated by both Grad-CAM++ and Grad-CAM on the Pascal VOC 2007 validation set (2510 images) (``incr'' denotes increase). The explanations were generated for decisions taken by the \textbf{AlexNet} architecture.\textit{(See Section \ref{Section:Objective evaluation of performance based on object recognition} for details of the metrics used.)}}
\label{supplementary alexnet pascal}
\vspace{-4mm}\end{table}
\begin{table}[h]
\begin{center}
\begin{tabular}{|l|c|c|}
\hline
Method & \textbf{Grad-CAM++} & Grad-CAM   \\
\hline\hline
- Average drop\% & \textbf{28.90} & 30.36\\
\textit{(Lower is better)} &  & \\
\hline
- \% incr. in confidence & \textbf{22.16} & 22.11 \\
\textit{(Higher is better)} &  & \\
\hline
- Win\% & \textbf{60.51} & 39.49\\
\textit{(Higher is better)} &  & \\
\hline
\end{tabular}
\end{center}
\caption{Results for objective evaluation of the explanations generated by both Grad-CAM++ and Grad-CAM on the ImageNet (ILSVRC2012) validation set (``incr'' denotes increase). The explanations were generated for decisions taken by the \textbf{ResNet-50} architecture.\textit{(See Section \ref{Section:Objective evaluation of performance based on object recognition} for details of the metrics used.)}}
\label{supplementary resnet imagenet}
\vspace{-4mm}\end{table}
\begin{table}
\begin{center}
\begin{tabular}{|l|c|c|}
\hline
Method & \textbf{Grad-CAM++} & Grad-CAM   \\
\hline\hline
- Average drop\% & \textbf{16.19} & 20.86\\
\textit{(Lower is better)} &  & \\
\hline
- \% incr. in confidence & 19.52 &  \textbf{21.99}\\
\textit{(Higher is better)} &  & \\
\hline
- Win\% & \textbf{58.61} & 41.39\\
\textit{(Higher is better)} &  & \\
\hline
\end{tabular}
\end{center}
\caption{Results for objective evaluation of the explanations generated by both Grad-CAM++ and Grad-CAM on the Pascal VOC 2007 validation set (2510 images) (``incr'' denotes increase). The explanations were generated for decisions taken by the \textbf{ResNet-50} architecture.\textit{(See Section \ref{Section:Objective evaluation of performance based on object recognition} for details of the metrics used.)}}
\label{supplementary resnet pascal}
\vspace{-4mm}\end{table}

\subsection{Additional Results on Image Captioning}
\label{Supplementary:B}
\begin{figure*}
\begin{center}
   \includegraphics[scale=1.2]{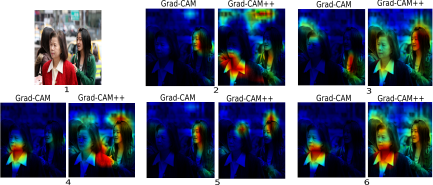}
\end{center}
   \caption{Visual explanations generated by Grad-CAM and Grad-CAM++ on a given image (1) predicting 5 different captions: (2) A woman in a red sweater and a girl is in front of a girl in a green hoodie and a girl with a brown jacket and pink purse; (3) A woman in a red sweater walks by two younger women talking near a busy street; (4) Two Asian women are talking behind an older woman who is wearing a red sweater; (5) A woman in a red sweater walks past two younger women who are chatting; and (6) Three asian women, two young, one old, on an urban sidewalk.}
\label{fig:caption result 3}
\vspace{-5mm}\end{figure*}

\begin{figure*}
\begin{center}
   \includegraphics[scale=1.2]{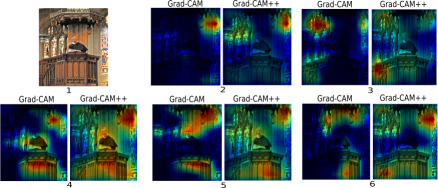}
\end{center}
   \caption{Visual explanations generated by Grad-CAM and Grad-CAM++ on a given image (1) predicting 5 different captions: (2) A priest stands in a pulpit giving a ceremony motioning with his hands in front of stained glass windows in the church; (3) A religious man giving a sermon at a beautifully carved pulpit with stained glass murals behind him; (4) A priest speaks from an ornate pulpit with stained glass pictures in the background; (5) A man is speaking at a podium in a church; and (6) A priest delivering mass in a church.}
\label{fig:caption result 4}
\vspace{-5mm}\end{figure*}

In continuation to the results in Section \ref{Section:Image Captioning}, we present here the additional results of the visual explanations on the considered images with five different captions in Figures \ref{fig:caption result 1}, \ref{fig:caption result 3} and \ref{fig:caption result 4}. We note that Grad-CAM++ provides more complete explanations in each of these images, as before.
\begin{figure*}[t]
\begin{center}
   \includegraphics[scale=1.2]{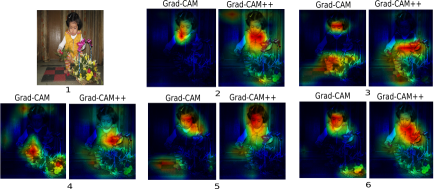}
\end{center}
   \caption{Visual explanations generated by Grad-CAM and Grad-CAM++ on a given image (1) predicting 5 different captions: (2) A little girl, with dark hair and a yellow vest with striped pants on, is crouching down next to a flower basket; (3) A small child wearing yellow and white is crouched by a basket holding a flower; (4) A young girl crouched on the floor picks at flowers in a basket; (5) One little girl in a yellow shirt carrying a basket of flowers; and (6) Little girl is looking at the flowers.}
\label{fig:caption result 2}
\vspace{-5mm}\end{figure*}


\ifCLASSOPTIONcaptionsoff
  \newpage
\fi

\end{document}